\documentclass[11pt]{article}
\usepackage{hyperref}
\usepackage{graphicx}
\usepackage{amssymb}
\usepackage[fleqn]{amsmath}
\usepackage{epstopdf}
\newcommand{\cut}[1]{}

\allowdisplaybreaks

\hypersetup{
    colorlinks=true,
    linkcolor=blue,
    urlcolor=blue
}

\title{The Matrix Calculus You Need For Deep Learning}
\author{
\href{http://explained.ai}{Terence Parr} and \href{http://www.fast.ai/about/\#jeremy}{Jeremy Howard}
}
\begin{document}
\maketitle

(We teach in University of San Francisco's \href{https://www.usfca.edu/arts-sciences/graduate-programs/data-science}{MS in Data Science program} and have other nefarious projects underway. You might know Terence as the creator of the \href{http://www.antlr.org}{ANTLR parser generator}. For more material, see Jeremy's \href{http://course.fast.ai}{fast.ai courses} and University of San Francisco's Data Institute \href{https://www.usfca.edu/data-institute/certificates/deep-learning-part-one}{in-person version of the deep learning course}.)

\href{http://explained.ai/matrix-calculus/index.html}{HTML version} (The PDF and HTML were generated from markup using \href{https://github.com/parrt/bookish}{bookish})

\begin{abstract}

This paper is an attempt to explain all the matrix calculus you need in order to understand the training of deep neural networks. We assume no math knowledge beyond what you learned in calculus 1, and provide links to help you refresh the necessary math where needed. Note that you do {\bf not} need to understand this material before you start learning to train and use deep learning in practice; rather, this material is for those who are already familiar with the basics of neural networks, and wish to deepen their understanding of the underlying math. Don't worry if you get stuck at some point along the way---just go back and reread the previous section, and try writing down and working through some examples. And if you're still stuck, we're happy to answer your questions in the \href{http://forums.fast.ai/c/theory}{Theory category at forums.fast.ai}. {\bf Note}: There is a \href{\#reference}{reference section} at the end of the paper summarizing all the key matrix calculus rules and terminology discussed here.

\end{abstract}

\pagebreak
{\small \setlength{\parskip}{0pt} \tableofcontents}
\pagebreak

\section{Introduction}\label{intro}

Most of us last saw calculus in school, but derivatives are a critical part of machine learning, particularly deep neural networks, which are trained by optimizing a loss function. Pick up a machine learning paper or the documentation of a library such as \href{http://pytorch.org}{PyTorch} and calculus comes screeching back into your life like distant relatives around the holidays.  And it's not just any old scalar calculus that pops up---you need differential {\em matrix calculus}, the shotgun wedding of \href{https://en.wikipedia.org/wiki/Linear\_algebra}{linear algebra} and \href{https://en.wikipedia.org/wiki/Multivariable\_calculus}{multivariate calculus}. 

Well... maybe {\em need} isn't the right word; Jeremy's courses show how to become a world-class deep learning practitioner with only a minimal level of scalar calculus, thanks to leveraging the automatic differentiation built in to modern deep learning libraries. But if you really want to really understand what's going on under the hood of these libraries, and grok academic papers discussing the latest advances in model training techniques, you'll need to understand certain bits of the field of matrix calculus.

For example, the activation of a single computation unit in a neural network is typically calculated using the dot product (from linear algebra) of an edge weight vector $\mathbf{w}$ with an input vector $\mathbf{x}$ plus a scalar bias (threshold): $z(\mathbf{x}) = \sum_i^n w_i x_i + b = \mathbf{w} \cdot \mathbf{x} + b$. Function $z(\mathbf{x})$ is called the unit's {\em affine function} and is followed by a \href{https://goo.gl/7BXceK}{rectified linear unit}, which clips negative values to zero: $max(0, z(\mathbf{x}))$. Such a computational unit is sometimes referred to as an ``artificial neuron'' and looks like:

\begin{center}
\includegraphics{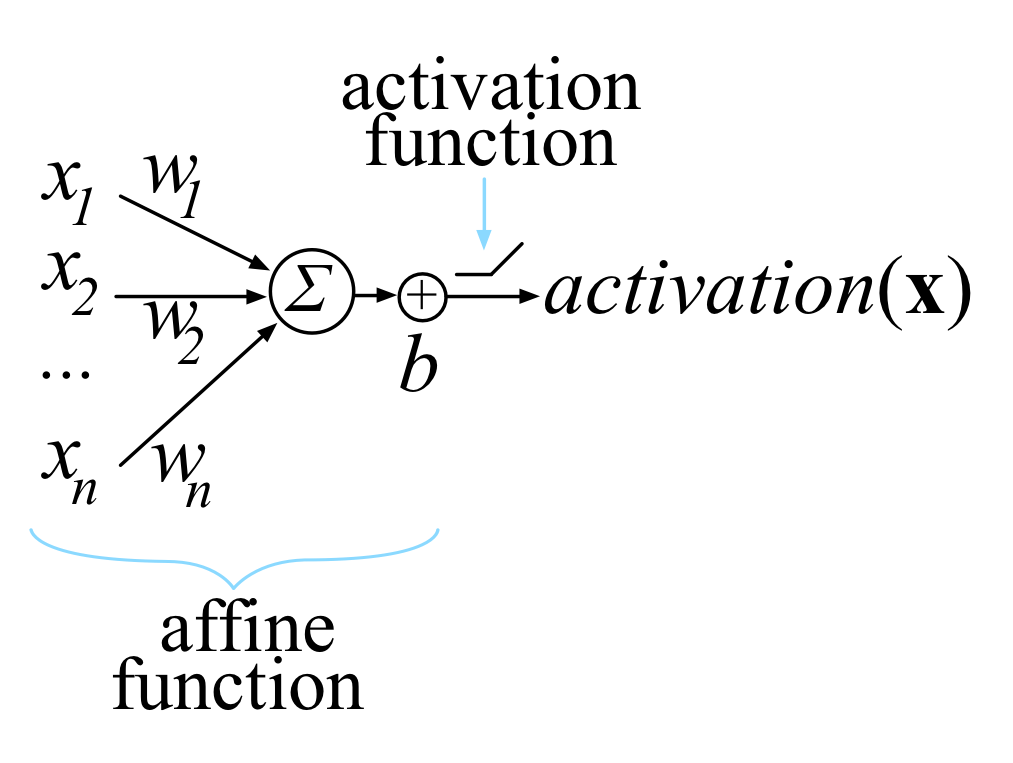}
\end{center}

Neural networks consist of many of these units, organized into multiple collections of neurons called {\em layers}. The activation of one layer's units become the input to the next layer's units. The activation of the unit or units in the final layer is called the network output.

{\em Training} this neuron means choosing weights $\mathbf{w}$ and bias $b$ so that we get the desired output for all $N$ inputs $\mathbf{x}$.  To do that, we minimize a {\em loss function} that compares the network's final $activation({\mathbf{x}})$ with the $target(\mathbf{x})$ (desired output of $\mathbf{x}$) for all input $\mathbf{x}$ vectors. To minimize the loss, we use some variation on gradient descent, such as plain \href{https://en.wikipedia.org/wiki/Stochastic\_gradient\_descent}{stochastic gradient descent} (SGD), SGD with momentum, or \href{https://en.wikipedia.org/wiki/Stochastic\_gradient\_descent\#Adam}{Adam}.   All of those require the partial derivative (the gradient) of $activation({\mathbf{x}})$ with respect to the model parameters $\mathbf{w}$ and $b$. Our goal is to gradually tweak $\mathbf{w}$ and $b$ so that the overall loss function keeps getting smaller across all $\mathbf{x}$ inputs.

If we're careful, we can derive the gradient by differentiating the scalar version of a common loss function (mean squared error):
\[
\frac{1}{N} \sum_{\mathbf{x}} (target(\mathbf{x}) - activation(\mathbf{x}))^2 = \frac{1}{N} \sum_{\mathbf{x}} (target(\mathbf{x}) - max(0, \sum_i^{|x|} w_i x_i + b))^2
\]
But this is just one neuron, and neural networks must train the weights and biases of all neurons in all layers simultaneously.  Because there are multiple inputs and (potentially) multiple network outputs, we really need general rules for the derivative of a function with respect to a vector and even rules for the derivative of a vector-valued function with respect to a vector.

This article walks through the derivation of some important rules for computing partial derivatives with respect to vectors, particularly those useful for training neural networks. This field is known as {\em matrix calculus}, and the good news is, we only need a small subset of that field, which we introduce here.  While there is a lot of online material on multivariate calculus and linear algebra, they are typically taught as two separate undergraduate courses so most material treats them in isolation.  The pages that do discuss matrix calculus often are really just lists of rules with minimal explanation or are just pieces of the story. They also tend to be quite obscure to all but a narrow audience of mathematicians, thanks to their use of dense notation and minimal discussion of foundational concepts. (See the annotated list of resources at the end.)

In contrast, we're going to rederive and rediscover some key matrix calculus rules in an effort to explain them. It turns out that matrix calculus is really not that hard! There aren't dozens of new rules to learn; just a couple of key concepts.  Our hope is that this short paper will get you started quickly in the world of matrix calculus as it relates to training neural networks. We're assuming you're already familiar with the basics of neural network architecture and training. If you're not, head over to \href{http://course.fast.ai}{Jeremy's course} and complete part 1 of that, then we'll see you back here when you're done. (Note that, unlike many more academic approaches, we strongly suggest {\em first} learning to train and use neural networks in practice and {\em then} study the underlying math. The math will be much more understandable with the context in place; besides, it's not necessary to grok all this calculus to become an effective practitioner.)

{\em A note on notation}: Jeremy's course exclusively uses code, instead of math notation, to explain concepts since unfamiliar functions in code are easy to search for and experiment with. In this paper, we do the opposite: there is a lot of math notation because one of the goals of this paper is to help you understand the notation that you'll see in deep learning papers and books. At the \href{\#notation}{end of the paper}, you'll find a brief table of the notation used, including a word or phrase you can use to search for more details.

\section{Review: Scalar derivative rules}\label{sec2}

Hopefully you remember some of these main scalar derivative rules. If your memory is a bit fuzzy on this, have a look at \href{https://www.khanacademy.org/math/ap-calculus-ab/ab-derivative-rules}{Khan academy vid on scalar derivative rules}.

\begin{tabular}{p{0.20\linewidth}lp{0.30\linewidth}l}
{\bf Rule}&{\bf $f(x)$}&{\bf Scalar derivative notation with respect to $x$}&{\bf Example}\\

{\bf Constant}&$c$&$0$&$\frac{d}{dx}99 = 0$\\
{\bf Multiplication by constant}&$cf$&$c \frac{df}{dx}$&$\frac{d}{dx}3x = 3$\\
{\bf Power Rule}&$x^n$&$nx^{n-1}$&$\frac{d}{dx}x^3 = 3x^2$\\
{\bf Sum Rule}&$f + g$&$\frac{df}{dx} + \frac{dg}{dx}$&$\frac{d}{dx} (x^2 + 3x) = 2x + 3$\\
{\bf Difference Rule}&$f - g$&$\frac{df}{dx} - \frac{dg}{dx}$&$\frac{d}{dx}(x^2 - 3x) = 2x - 3$\\
{\bf Product Rule}&$fg$&$f \frac{dg}{dx} + \frac{df}{dx} g$&$\frac{d}{dx}x^2x = x^2 + x2x = 3x^2$\\
{\bf Chain Rule}&$f(g(x))$&$\frac{df(u)}{du}\frac{du}{dx}$,  let $u=g(x)$&$\frac{d}{dx} ln(x^2) = \frac{1}{x^2}2x = \frac{2}{x}$\\

\end{tabular}

There are other rules for trigonometry, exponentials, etc., which you can find at \href{https://www.khanacademy.org/math/differential-calculus}{Khan Academy differential calculus course}.

When a function has a single parameter, $f(x)$, you'll often see $f'$ and $f'(x)$ used as shorthands for $\frac{d}{dx}f(x)$. We recommend against this notation as it does not make clear the variable we're taking the derivative with respect to. 

You can think of $\frac{d}{dx}$ as an operator that maps a function of one parameter to another function.  That means that $\frac{d}{dx} f(x)$ maps $f(x)$ to its derivative with respect to $x$, which is the same thing as $\frac{df(x)}{dx}$. Also, if $y = f(x)$, then $\frac{dy}{dx} = \frac{df(x)}{dx} = \frac{d}{dx}f(x)$. Thinking of the derivative as an operator helps to simplify complicated derivatives because the operator is distributive and lets us pull out constants. For example, in the following equation, we can pull out the constant 9 and distribute the derivative operator across the elements within the parentheses.
\[
\frac{d}{dx} 9(x + x^2) = 9 \frac{d}{dx}(x + x^2) = 9 (\frac{d}{dx}x + \frac{d}{dx}x^2) = 9(1 + 2x) = 9 + 18x
\]
That procedure reduced the derivative of $9(x + x^2)$ to a bit of arithmetic and the derivatives of $x$ and $x^2$, which are much easier to solve than the original derivative.

\section{Introduction to vector calculus and partial derivatives}\label{sec3}

Neural network layers are not single functions of a single parameter, $f(x)$. So, let's move on to functions of multiple parameters such as $f(x,y)$. For example, what is the derivative of $xy$ (i.e., the multiplication of $x$ and $y$)? In other words, how does the product $xy$ change when we wiggle the variables? Well, it depends on whether we are changing $x$ or $y$.  We compute derivatives with respect to one variable (parameter) at a time, giving us two different {\em partial derivatives} for this two-parameter function (one for $x$ and one for $y$).  Instead of using operator $\frac{d}{dx}$, the partial derivative operator is  $\frac{\partial}{\partial x}$ (a stylized $d$ and not the Greek letter $\delta$). So, $\frac{\partial }{\partial x}xy$ and $\frac{\partial }{\partial y}xy$ are the partial derivatives of $xy$; often, these are just called the {\em partials}.  For functions of a single parameter, operator $\frac{\partial}{\partial x}$ is equivalent to $\frac{d}{dx}$ (for sufficiently smooth functions). However, it's better to use $\frac{d}{dx}$ to make it clear you're referring to a scalar derivative.

The partial derivative with respect to $x$ is just the usual scalar derivative, simply treating any other variable in the equation as a constant.  Consider function $f(x,y) = 3x^2y$. The partial derivative with respect to $x$ is written $\frac{\partial}{\partial x} 3x^2y$. There are three constants from the perspective of $\frac{\partial}{\partial x}$: 3, 2, and $y$. Therefore, $\frac{\partial}{\partial x} 3yx^2 = 3y\frac{\partial}{\partial x} x^2 = 3y2x = 6yx$. The partial derivative with respect to $y$ treats $x$ like a constant: $\frac{\partial}{\partial y} 3x^2y = 3x^2\frac{\partial}{\partial y} y = 3x^2\frac{\partial y}{\partial y} = 3x^2 \times 1 = 3x^2$.  It's a good idea to derive these yourself before continuing otherwise the rest of the article won't make sense.  Here's the \href{https://www.khanacademy.org/math/multivariable-calculus/multivariable-derivatives/partial-derivative-and-gradient-articles/a/introduction-to-partial-derivatives}{Khan Academy video on partials} if you need help.

To make it clear we are doing vector calculus and not just multivariate calculus, let's consider what we do with the partial derivatives $\frac{\partial f(x,y)}{\partial x}$ and $\frac{\partial f(x,y)}{\partial y}$ (another way to say $\frac{\partial}{\partial x}f(x,y)$ and $\frac{\partial }{\partial y}f(x,y)$) that we computed for $f(x,y) = 3x^2y$.  Instead of having them just floating around and not organized in any way, let's organize them into a horizontal vector. We call this vector the {\em gradient} of $f(x,y)$ and write it as:
\[\nabla f(x,y)  = [ \frac{\partial f(x,y)}{\partial x}, \frac{\partial f(x,y)}{\partial y}] = [6yx, 3x^2]\]
So the gradient of $f(x,y)$ is simply a vector of its partials. Gradients are part of the vector calculus world, which deals with functions that map $n$ scalar parameters to a single scalar.  Now, let's get crazy and consider derivatives of multiple functions simultaneously.

\section{Matrix calculus}\label{sec4}

When we move from derivatives of one function to derivatives of many functions, we move from the world of vector calculus to matrix calculus. Let's compute partial derivatives for two functions, both of which take two parameters.  We can keep the same $f(x,y) = 3x^2y$ from the last section, but let's also bring in $g(x,y) = 2x + y^8$.  The gradient for $g$ has two entries, a partial derivative for each parameter:
\[\frac{\partial g(x,y)}{\partial x} = \frac{\partial 2x}{\partial x} + \frac{\partial y^8}{\partial x} = 2\frac{\partial x}{\partial x} + 0 = 2 \times 1 = 2\]
and
\[\frac{\partial g(x,y)}{\partial y} = \frac{\partial 2x}{\partial y} + \frac{\partial y^8}{\partial y} = 0 + 8y^7 = 8y^7\]
giving us gradient $\nabla g(x,y) = [2, 8y^7]$.

Gradient vectors organize all of the partial derivatives for a specific scalar function. If we have two functions, we can also organize their gradients into a matrix by stacking the gradients. When we do so, we get the {\em Jacobian matrix} (or just the {\em Jacobian}) where the gradients are rows:
\[J =
\begin{bmatrix}
	\nabla f(x,y)\\
	\nabla g(x,y)
\end{bmatrix} = \begin{bmatrix}
 \frac{\partial f(x,y)}{\partial x} & \frac{\partial f(x,y)}{\partial y}\\
 \frac{\partial g(x,y)}{\partial x} & \frac{\partial g(x,y)}{\partial y}\\
\end{bmatrix} = \begin{bmatrix}
	6yx & 3x^2\\
	2 & 8y^7
\end{bmatrix}\]
Welcome to matrix calculus!

{\bf Note that there are multiple ways to represent the Jacobian.} We are using the so-called \href{https://en.wikipedia.org/wiki/Matrix\_calculus\#Layout\_conventions}{numerator layout} but many papers and software will use the {\em denominator layout}. This is just transpose of the numerator layout Jacobian (flip it around its diagonal):
\[
\begin{bmatrix}
	6yx & 2\\
	3x^2 & 8y^7
\end{bmatrix}
\]

\subsection{Generalization of the Jacobian}\label{sec4.1}

So far, we've looked at a specific example of a Jacobian matrix. To define the Jacobian matrix more generally, let's combine multiple parameters into a single vector argument: $f(x,y,z) \Rightarrow f(\mathbf{x})$. (You will sometimes see notation $\vec{x}$  for vectors in the literature as well.) Lowercase letters in bold font such as $\mathbf{x}$ are vectors and those in italics font like $x$ are scalars. $x_i$ is the $i^{th}$ element of vector $\mathbf{x}$ and is in italics because a single vector element is a scalar. We also have to define an orientation for vector $\mathbf{x}$. We'll assume that all vectors are vertical by default of size $n \times 1$:
\[\mathbf{x} = \begin{bmatrix}
           x_1\\
           x_2\\
           \vdots \\
           x_n\\
           \end{bmatrix}\]
With multiple scalar-valued functions, we can combine them all into a vector just like we did with the parameters. Let $\mathbf{y} = \mathbf{f}(\mathbf{x})$ be a vector of $m$ scalar-valued functions that each take a vector $\mathbf{x}$ of length $n=|\mathbf{x}|$ where $|\mathbf{x}|$ is the cardinality (count) of elements in $\mathbf{x}$. Each $f_i$ function within $\mathbf{f}$ returns a scalar just as in the previous section:
\[
\begin{array}{lcl}
 y_1 & = & f_1(\mathbf{x})\\
 y_2 & = & f_2(\mathbf{x})\\
 & \vdots & \\
 y_m & = & f_m(\mathbf{x})\\
\end{array}\]
For instance, we'd represent $f(x,y) = 3x^2y$ and $g(x,y) = 2x + y^8$ from the last section as
\[
\begin{array}{lllllllll}
 y_1 = f_1(\mathbf{x}) = 3x_1^2x_2  &&&(\text{substituting }x_1 \text{ for }x, x_2 \text{ for }y)\\
 y_2 = f_2(\mathbf{x}) = 2x_1 + x_2^8
\end{array}
\]
It's very often the case that $m=n$ because we will have a scalar function result for each element of the $\mathbf{x}$ vector.  For example, consider the identity function $\mathbf{y} = \mathbf{f(x)} = \mathbf{x}$:
\[\begin{array}{lclcc}
 y_1 & = & f_1(\mathbf{x})& = & x_1\\
 y_2 & = & f_2(\mathbf{x})& = & x_2\\
 & \vdots & \\
 y_n & = & f_n(\mathbf{x})& = & x_n\\
\end{array}\]
So we have $m=n$ functions and parameters, in this case. Generally speaking, though, the Jacobian matrix is the collection of all $m \times n$ possible partial derivatives ($m$ rows and $n$ columns), which is the stack of $m$ gradients with respect to $\mathbf{x}$:
\[
\frac{\partial \mathbf{y}}{\partial \mathbf{x}} = \begin{bmatrix}
\nabla f_1(\mathbf{x}) \\
\nabla f_2(\mathbf{x})\\
\ldots\\
\nabla f_m(\mathbf{x})
\end{bmatrix} = \begin{bmatrix}
\frac{\partial}{\partial \mathbf{x}} f_1(\mathbf{x}) \\
\frac{\partial}{\partial \mathbf{x}} f_2(\mathbf{x})\\
\ldots\\
\frac{\partial}{\partial \mathbf{x}} f_m(\mathbf{x})
\end{bmatrix} = \begin{bmatrix}
\frac{\partial}{\partial {x_1}} f_1(\mathbf{x})~ \frac{\partial}{\partial {x_2}} f_1(\mathbf{x}) ~\ldots~ \frac{\partial}{\partial {x_n}} f_1(\mathbf{x}) \\
\frac{\partial}{\partial {x_1}} f_2(\mathbf{x})~ \frac{\partial}{\partial {x_2}} f_2(\mathbf{x}) ~\ldots~ \frac{\partial}{\partial {x_n}} f_2(\mathbf{x}) \\
\ldots\\
~\frac{\partial}{\partial {x_1}} f_m(\mathbf{x})~ \frac{\partial}{\partial {x_2}} f_m(\mathbf{x}) ~\ldots~ \frac{\partial}{\partial {x_n}} f_m(\mathbf{x}) \\
\end{bmatrix}
\]
Each $\frac{\partial}{\partial \mathbf{x}} f_i(\mathbf{x})$ is a horizontal $n$-vector because the partial derivative is with respect to a vector, $\mathbf{x}$, whose length is $n = |\mathbf{x}|$.  The width of the Jacobian is $n$ if we're taking the partial derivative with respect to $\mathbf{x}$ because there are $n$ parameters we can wiggle, each potentially changing the function's value. Therefore, the Jacobian is always $m$ rows for $m$ equations.  It helps to think about the possible Jacobian shapes visually:
\begin{center}

\begin{tabular}{c|ccl}
  & \begin{tabular}[t]{c}
  scalar\\
  \framebox(18,18){$x$}\\
  \end{tabular} & \begin{tabular}{c}
  vector\\
  \framebox(18,40){$\mathbf{x}$}
  \end{tabular}\\
\hline
\\[\dimexpr-\normalbaselineskip+5pt]
\begin{tabular}[b]{c}
  scalar\\
  \framebox(18,18){$f$}\\
  \end{tabular} &\framebox(18,18){$\frac{\partial f}{\partial {x}}$} & \framebox(40,18){$\frac{\partial f}{\partial {\mathbf{x}}}$}&\\
\begin{tabular}[b]{c}
  vector\\
  \framebox(18,40){$\mathbf{f}$}\\
  \end{tabular} & \framebox(18,40){$\frac{\partial \mathbf{f}}{\partial {x}}$} & \framebox(40,40){$\frac{\partial \mathbf{f}}{\partial \mathbf{x}}$}\\
\end{tabular}

\end{center}
The Jacobian of the identity function $\mathbf{f(x)} = \mathbf{x}$, with $f_i(\mathbf{x}) = x_i$, has $n$ functions and each function has $n$ parameters held in a single vector $\mathbf{x}$. The Jacobian is, therefore, a square matrix since $m=n$:
\begin{center}

\begin{eqnarray*}
	\frac{\partial \mathbf{y}}{\partial \mathbf{x}} = \begin{bmatrix}
	\frac{\partial}{\partial \mathbf{x}} f_1(\mathbf{x}) \\
	\frac{\partial}{\partial \mathbf{x}} f_2(\mathbf{x})\\
	\ldots\\
	\frac{\partial}{\partial \mathbf{x}} f_m(\mathbf{x})
	\end{bmatrix} &=& \begin{bmatrix}
	\frac{\partial}{\partial {x_1}} f_1(\mathbf{x})~ \frac{\partial}{\partial {x_2}} f_1(\mathbf{x}) ~\ldots~  \frac{\partial}{\partial {x_n}} f_1(\mathbf{x}) \\
	\frac{\partial}{\partial {x_1}} f_2(\mathbf{x})~ \frac{\partial}{\partial {x_2}} f_2(\mathbf{x}) ~\ldots~  \frac{\partial}{\partial {x_n}} f_2(\mathbf{x}) \\
	\ldots\\
	~\frac{\partial}{\partial {x_1}} f_m(\mathbf{x})~ \frac{\partial}{\partial {x_2}} f_m(\mathbf{x}) ~\ldots~ \frac{\partial}{\partial {x_n}} f_m(\mathbf{x}) \\
	\end{bmatrix}\\\\
	& = & \begin{bmatrix}
	\frac{\partial}{\partial {x_1}} x_1~ \frac{\partial}{\partial {x_2}} x_1 ~\ldots~ \frac{\partial}{\partial {x_n}} x_1 \\
	\frac{\partial}{\partial {x_1}} x_2~ \frac{\partial}{\partial {x_2}} x_2 ~\ldots~ \frac{\partial}{\partial {x_n}} x_2 \\
	\ldots\\
	~\frac{\partial}{\partial {x_1}} x_n~ \frac{\partial}{\partial {x_2}} x_n ~\ldots~ \frac{\partial}{\partial {x_n}} x_n \\
	\end{bmatrix}\\\\
	& & (\text{and since } \frac{\partial}{\partial {x_j}} x_i = 0 \text{ for } j \neq i)\\
	 & = & \begin{bmatrix}
	\frac{\partial}{\partial {x_1}} x_1 & 0 & \ldots& 0 \\
	0 & \frac{\partial}{\partial {x_2}} x_2 &\ldots & 0 \\
	& & \ddots\\
	0 & 0 &\ldots& \frac{\partial}{\partial {x_n}} x_n \\
	\end{bmatrix}\\\\
	 & = & \begin{bmatrix}
	1 & 0 & \ldots& 0 \\
	0 &1 &\ldots & 0 \\
	& & \ddots\\
	0 & 0 & \ldots &1 \\
	\end{bmatrix}\\\\
	& = & I ~~~(I \text{ is the identity matrix with ones down the diagonal})\\
	\end{eqnarray*}

\end{center}
Make sure that you can derive each step above before moving on. If you get stuck, just consider each element of the matrix in isolation and apply the usual scalar derivative rules.   That is a generally useful trick: Reduce vector expressions down to a set of scalar expressions and then take all of the partials, combining the results appropriately into vectors and matrices at the end.

Also be careful to track whether a matrix is vertical, $\mathbf{x}$, or horizontal, $\mathbf{x}^T$ where $\mathbf{x}^T$ means $\mathbf{x}$ transpose. Also make sure you pay attention to whether something is a scalar-valued function, $y = ...\,$, or a vector of functions (or a vector-valued function), $\mathbf{y} = ...\,$.

\subsection{Derivatives of vector element-wise binary operators}\label{sec4.2}

Element-wise binary operations on vectors, such as vector addition $\mathbf{w} + \mathbf{x}$, are important because we can express many common vector operations, such as the multiplication of a vector by a scalar, as element-wise binary operations.  By ``element-wise binary operations'' we simply mean applying an operator to the first item of each vector to get the first item of the output, then to the second items of the inputs for the second item of the output, and so forth. This is how all the basic math operators are applied by default in numpy or tensorflow, for example.  Examples that often crop up in deep learning are $max(\mathbf{w},\mathbf{x})$ and $\mathbf{w} > \mathbf{x}$ (returns a vector of ones and zeros). 

We can generalize the element-wise binary operations with notation $\mathbf{y} = \mathbf{f(w)} \bigcirc \mathbf{g(x)}$ where $m=n=|y|=|w|=|x|$. (Reminder: $|x|$ is the number of items in $x$.) The $\bigcirc$ symbol represents any element-wise operator (such as $+$) and not the $\circ$ function composition operator.  Here's what equation $\mathbf{y} = \mathbf{f(w)} \bigcirc \mathbf{g(x)}$ looks like when we zoom in to examine the scalar equations:
\[\begin{bmatrix}
           y_1\\
           y_2\\
           \vdots \\
           y_n\\
           \end{bmatrix} = \begin{bmatrix}
           f_{1}(\mathbf{w}) \bigcirc g_{1}(\mathbf{x})\\
           f_{2}(\mathbf{w}) \bigcirc g_{2}(\mathbf{x})\\
           \vdots \\
           f_{n}(\mathbf{w}) \bigcirc g_{n}(\mathbf{x})\\
         \end{bmatrix}\]
where we write $n$ (not $m$) equations vertically to emphasize the fact that the result of element-wise operators give $m=n$ sized vector results.

Using the ideas from the last section, we can see that the general case for the Jacobian with respect to $\mathbf{w}$ is the square matrix:
\[J_\mathbf{w} = 
\frac{\partial \mathbf{y}}{\partial \mathbf{w}}  = \begin{bmatrix}
\frac{\partial}{\partial w_1} ( f_{1}(\mathbf{w}) \bigcirc g_{1}(\mathbf{x}) ) & \frac{\partial}{\partial w_2} ( f_{1}(\mathbf{w}) \bigcirc g_{1}(\mathbf{x}) ) & \ldots & \frac{\partial}{\partial w_n} ( f_{1}(\mathbf{w}) \bigcirc g_{1}(\mathbf{x}) )\\
\frac{\partial}{\partial w_1} ( f_{2}(\mathbf{w}) \bigcirc g_{2}(\mathbf{x}) ) & \frac{\partial}{\partial w_2} ( f_{2}(\mathbf{w}) \bigcirc g_{2}(\mathbf{x}) ) & \ldots & \frac{\partial}{\partial w_n} ( f_{2}(\mathbf{w}) \bigcirc g_{2}(\mathbf{x}) )\\
& \ldots\\
\frac{\partial}{\partial w_1} ( f_{n}(\mathbf{w}) \bigcirc g_{n}(\mathbf{x}) ) & \frac{\partial}{\partial w_2} ( f_{n}(\mathbf{w}) \bigcirc g_{n}(\mathbf{x}) ) & \ldots & \frac{\partial}{\partial w_n} ( f_{n}(\mathbf{w}) \bigcirc g_{n}(\mathbf{x}) )\\
\end{bmatrix}\]
and the Jacobian with respect to $\mathbf{x}$ is:
\[J_\mathbf{x} = 
\frac{\partial \mathbf{y}}{\partial \mathbf{x}}  = \begin{bmatrix}
\frac{\partial}{\partial x_1} ( f_{1}(\mathbf{w}) \bigcirc g_{1}(\mathbf{x}) ) & \frac{\partial}{\partial x_2} ( f_{1}(\mathbf{w}) \bigcirc g_{1}(\mathbf{x}) ) & \ldots & \frac{\partial}{\partial x_n} ( f_{1}(\mathbf{w}) \bigcirc g_{1}(\mathbf{x}) )\\
\frac{\partial}{\partial x_1} ( f_{2}(\mathbf{w}) \bigcirc g_{2}(\mathbf{x}) ) & \frac{\partial}{\partial x_2} ( f_{2}(\mathbf{w}) \bigcirc g_{2}(\mathbf{x}) ) & \ldots & \frac{\partial}{\partial x_n} ( f_{2}(\mathbf{w}) \bigcirc g_{2}(\mathbf{x}) )\\
& \ldots\\
\frac{\partial}{\partial x_1} ( f_{n}(\mathbf{w}) \bigcirc g_{n}(\mathbf{x}) ) & \frac{\partial}{\partial x_2} ( f_{n}(\mathbf{w}) \bigcirc g_{n}(\mathbf{x}) ) & \ldots & \frac{\partial}{\partial x_n} ( f_{n}(\mathbf{w}) \bigcirc g_{n}(\mathbf{x}) )\\
\end{bmatrix}\]
That's quite a furball, but fortunately the Jacobian is very often a diagonal matrix, a matrix that is zero everywhere but the diagonal. Because this greatly simplifies the Jacobian, let's examine in detail when the Jacobian reduces to a diagonal matrix for element-wise operations. 

In a diagonal Jacobian, all elements off the diagonal are zero, $\frac{\partial}{\partial w_j} ( f_i(\mathbf{w}) \bigcirc g_i(\mathbf{x}) ) = 0$ where $j \neq i$. (Notice that we are taking the partial derivative with respect to $w_j$ not $w_i$.) Under what conditions are those off-diagonal elements zero? Precisely when $f_i$ and $g_i$ are contants with respect to $w_j$, $\frac{\partial}{\partial w_j} f_i(\mathbf{w}) = \frac{\partial}{\partial w_j} g_i(\mathbf{x}) = 0$.  Regardless of the operator, if those partial derivatives go to zero, the operation goes to zero, $0 \bigcirc 0 = 0$ no matter what, and the partial derivative of a constant is zero.

Those partials go to zero when $f_i$ and $g_i$ are not functions of $w_j$. We know that element-wise operations imply that $f_i$ is purely a function of $w_i$ and $g_i$  is purely a function of $x_i$. For example, $\mathbf{w}+\mathbf{x}$ sums $w_i + x_i$. Consequently,  $f_i(\mathbf{w}) \bigcirc g_i(\mathbf{x})$ reduces to $f_i(w_i) \bigcirc g_i(x_i)$ and the goal becomes $\frac{\partial}{\partial w_j} f_i(w_i) = \frac{\partial}{\partial w_j} g_i(x_i) = 0$. $f_i(w_i)$ and $g_i(x_i)$ look like constants to the partial differentiation operator with respect to $w_j$ when $j \neq i$ so the partials are zero off the diagonal. (Notation $f_i(w_i)$ is technically an abuse of our notation because $f_i$ and $g_i$ are functions of vectors not individual elements. We should really write something like $\hat f_{i}(w_i) = f_{i}(\mathbf{w})$, but that would muddy the equations further, and programmers are comfortable overloading functions, so we'll proceed with the notation anyway.)  

We'll take advantage of this simplification later and refer to the constraint that $f_i(\mathbf{w})$ and $g_i(\mathbf{x})$ access at most $w_i$ and $x_i$, respectively, as the {\em element-wise diagonal condition}.

Under this condition, the elements along the diagonal of the Jacobian are $\frac{\partial}{\partial w_i} ( f_i(w_i) \bigcirc g_i(x_i) )$:
\[\frac{\partial \mathbf{y}}{\partial \mathbf{w}}  = \begin{bmatrix}
\frac{\partial}{\partial w_1} ( f_{1}(w_1) \bigcirc g_{1}(x_1) )\\
& \frac{\partial}{\partial w_2} (f_{2}(w_2) \bigcirc g_{2}(x_2) ) & & \text{\huge0}\\
& & \ldots \\
\text{\huge0}& & & \frac{\partial}{\partial w_n} (f_{n}(w_n) \bigcirc g_{n}(x_n) )
\end{bmatrix}\]
(The large ``0''s are a shorthand indicating all of the off-diagonal are 0.)

More succinctly, we can write:
\[\frac{\partial \mathbf{y}}{\partial \mathbf{w}} = diag \left( \frac{\partial}{\partial w_1}(f_{1}(w_1) \bigcirc g_{1}(x_1)),~ \frac{\partial}{\partial w_2}(f_{2}(w_2) \bigcirc g_{2}(x_2)),~ \ldots,~ \frac{\partial}{\partial w_n}(f_{n}(w_n) \bigcirc g_{n}(x_n)) \right)\]
and
\[\frac{\partial \mathbf{y}}{\partial \mathbf{x}} = diag \left( \frac{\partial}{\partial x_1}(f_{1}(w_1) \bigcirc g_{1}(x_1)),~ \frac{\partial}{\partial x_2}(f_{2}(w_2) \bigcirc g_{2}(x_2)),~ \ldots,~ \frac{\partial}{\partial x_n}(f_{n}(w_n) \bigcirc g_{n}(x_n)) \right)\]
where $diag(\mathbf{x})$ constructs a matrix whose diagonal elements are taken from vector $\mathbf{x}$. 

Because we do lots of simple vector arithmetic, the general function $\mathbf{f(w)}$ in the binary element-wise operation is often just the vector $\mathbf{w}$.  Any time the general function is a vector, we know that $f_i(\mathbf{w})$ reduces to $f_i(w_i) = w_i$. For example, vector addition $\mathbf{w + x}$ fits our element-wise diagonal condition because $\mathbf{f(w)} + \mathbf{g(x)}$ has scalar equations $y_i = f_i(\mathbf{w}) + g_i(\mathbf{x})$ that reduce to just $y_i = f_i(w_i) + g_i(x_i) = w_i + x_i$ with partial derivatives:
\[\frac{\partial}{\partial w_i} ( f_{i}(w_i) + g_{i}(x_i) ) = \frac{\partial}{\partial w_i}(w_i + x_i) = 1 + 0 = 1\]\[\frac{\partial}{\partial x_i} ( f_{i}(w_i) + g_{i}(x_i) ) = \frac{\partial}{\partial x_i}(w_i + x_i) = 0 + 1 = 1\]
That gives us $\frac{\partial (\mathbf{w+x})}{\partial \mathbf{w}} = \frac{\partial (\mathbf{w+x})}{\partial \mathbf{x}} = I$, the identity matrix, because every element along the diagonal is 1. $I$ represents the square identity matrix of appropriate dimensions that is zero everywhere but the diagonal, which contains all ones.  

Given the simplicity of this special case, $f_i(\mathbf{w})$ reducing to $f_i(w_i)$, you should be able to derive the Jacobians for the common element-wise binary operations on vectors:
\[
\begin{array}{lllllllll}
        \text{\bf{Op}} &  & {\text{\bf Partial with respect to }} \mathbf{w} \\
        + &  & \frac{\partial (\mathbf{w+x})}{\partial \mathbf{w}} = diag(\ldots \frac{\partial (w_i + x_i)}{\partial w_i} \ldots) = diag(\vec{1}) = I \\\\
        - &  & \frac{\partial (\mathbf{w-x})}{\partial \mathbf{w}}  =  diag(\ldots\frac{\partial (w_i - x_i)}{\partial w_i}\ldots) =  diag(\vec{1})  =  I \\\\
        \otimes &  & \frac{\partial (\mathbf{w \otimes x})}{\partial \mathbf{w}}  =  diag(\ldots\frac{\partial (w_i \times x_i)}{\partial w_i} \ldots)  =  diag(\mathbf{x}) \\\\
        \oslash &  & \frac{\partial (\mathbf{w \oslash x})}{\partial \mathbf{w}}  =  diag(\ldots\frac{\partial (w_i / x_i)}{\partial w_i}\ldots)  =  diag(\ldots \frac{1}{x_i} \ldots) \\\\
\end{array}
\]\[
\begin{array}{lllllllll}
        \text{\bf{Op}} &  &  {\text{\bf Partial with respect to }}\mathbf{x}\\
        + &  & \frac{\partial (\mathbf{w+x})}{\partial \mathbf{x}} =  I\\\\
        - &  & \frac{\partial (\mathbf{w-x})}{\partial \mathbf{x}}  =  diag(\ldots\frac{\partial (w_i - x_i)}{\partial x_i}\ldots)  =  diag(-\vec{1})  =  -I \\\\
        \otimes &  &  \frac{\partial (\mathbf{w \otimes x})}{\partial \mathbf{x}}  =  diag(\mathbf{w})\\\\
        \oslash &  &  \frac{\partial (\mathbf{w \oslash x})}{\partial \mathbf{x}}  =  diag(\ldots \frac{-w_i}{x_i^2} \ldots)\\
\end{array}
\]
The $\otimes$ and $\oslash$ operators are element-wise multiplication and division; $\otimes$ is sometimes called the {\em Hadamard product}. There isn't a standard notation for element-wise multiplication and division so we're using an approach consistent with our general binary operation notation.

\subsection{Derivatives involving scalar expansion}\label{sec4.3}

When we multiply or add scalars to vectors, we're implicitly expanding the scalar to a vector and then performing an element-wise binary operation. For example, adding scalar $z$  to vector $\mathbf{x}$, $\mathbf{y} = \mathbf{x} + z$, is really $\mathbf{y} = \mathbf{f(x)} + \mathbf{g}(z)$ where $\mathbf{f(x)} = \mathbf{x}$ and $\mathbf{g}(z) = \vec{1} z$. (The notation $\vec{1}$ represents a vector of ones of appropriate length.)  $z$ is any scalar that doesn't depend on $\mathbf{x}$, which is useful because then $\frac{\partial z}{\partial x_i} = 0$ for any $x_i$ and that will simplify our partial derivative computations. (It's okay to think of variable $z$ as a constant for our discussion here.)  Similarly, multiplying by a scalar, $\mathbf{y} = \mathbf{x} z$, is really $\mathbf{y} = \mathbf{f(x)} \otimes \mathbf{g}(z) = \mathbf{x} \otimes \vec{1}z$ where $\otimes$ is the element-wise  multiplication (Hadamard product) of the two vectors.

The partial derivatives of vector-scalar addition and multiplication with respect to vector $\mathbf{x}$ use our element-wise rule:
\[ \frac{\partial \mathbf{y}}{\partial \mathbf{x}} = diag \left( \ldots \frac{\partial}{\partial x_i} ( f_i(x_i) \bigcirc g_i(z) ) \ldots \right)\]
This follows because functions $\mathbf{f(x)} = \mathbf{x}$ and $\mathbf{g}(z) = \vec{1} z$ clearly satisfy our element-wise diagonal condition for the Jacobian (that $f_i(\mathbf{x})$ refer at most to $x_i$ and $g_i(z)$ refers to the $i^{th}$ value of the $\vec{1}z$ vector). 

Using the usual rules for scalar partial derivatives, we arrive at the following diagonal elements of the Jacobian for vector-scalar addition:
\[\frac{\partial}{\partial x_i} ( f_i(x_i) + g_i(z) ) = \frac{\partial (x_i + z)}{\partial x_i} = \frac{\partial x_i}{\partial x_i} + \frac{\partial z}{\partial x_i} = 1 + 0 = 1\]
So, $\frac{\partial}{\partial \mathbf{x}} ( \mathbf{x} + z ) = diag(\vec{1}) = I$.

Computing the partial derivative with respect to the scalar parameter $z$, however, results in a vertical vector, not a diagonal matrix. The elements of the vector are:
\[ \frac{\partial}{\partial z} ( f_i(x_i) + g_i(z) ) = \frac{\partial (x_i + z)}{\partial z} = \frac{\partial x_i}{\partial z} + \frac{\partial z}{\partial z} = 0 + 1 = 1\]
Therefore, $\frac{\partial}{\partial z} ( \mathbf{x} + z ) = \vec{1}$.

The diagonal elements of the Jacobian for vector-scalar multiplication involve the product rule for scalar derivatives:
\[ \frac{\partial}{\partial x_i} ( f_i(x_i) \otimes g_i(z) ) = x_i  \frac{\partial z}{\partial x_i} + z  \frac{\partial x_i}{\partial x_i} = 0 + z = z\]
So, $\frac{\partial}{\partial \mathbf{x}} ( \mathbf{x} z ) = diag(\vec{1}  z) = I z$. 

The partial derivative with respect to scalar parameter $z$ is a vertical vector whose elements are:
\[\frac{\partial}{\partial z} ( f_i(x_i) \otimes g_i(z) ) = x_i \frac{\partial z}{\partial z} + z \frac{\partial x_i}{\partial z} = x_i + 0 = x_i\]
This gives us $\frac{\partial}{\partial z} ( \mathbf{x} z ) = \mathbf{x}$.

\subsection{Vector sum reduction}\label{sec4.4}

Summing up the elements of a vector is an important operation in deep learning, such as the network loss function, but we can also use it as a way to simplify computing the derivative of vector dot product and other operations that reduce vectors to scalars.

Let $y = sum( \mathbf{f}(\mathbf{x})) = \sum_{i=1}^n f_i(\mathbf{x})$.  Notice we were careful here to leave the parameter as a vector $\mathbf{x}$ because each function $f_i$ could use all values in the vector, not just $x_i$. The sum is over the {\bf results} of the function and not the parameter. The gradient ($1 \times n$ Jacobian) of vector summation is:
\[
\begin{array}{lcllll}
 \frac{\partial y}{\partial \mathbf{x}} & = & \begin{bmatrix} \frac{\partial y}{\partial x_1}, \frac{\partial y}{\partial x_2}, \ldots, \frac{\partial y}{\partial x_n} \end{bmatrix}\\\\
  & = & \begin{bmatrix} \frac{\partial}{\partial x_1} \sum_i f_i(\mathbf{x}),~ \frac{\partial}{\partial x_2} \sum_i f_i(\mathbf{x}),~ \ldots,~ \frac{\partial}{\partial x_n} \sum_i  f_i(\mathbf{x}) \end{bmatrix} \\\\
 & = & \begin{bmatrix} \sum_i \frac{\partial f_i(\mathbf{x})}{\partial x_1},~ \sum_i \frac{\partial f_i(\mathbf{x})}{\partial x_2},~ \ldots,~ \sum_i \frac{\partial f_i(\mathbf{x})}{\partial x_n}  \end{bmatrix}&&&(\text{move derivative inside }\sum)\\
\end{array}
\]
(The summation inside the gradient elements can be tricky so make sure to keep your notation consistent.)

Let's look at the gradient of the simple $y = sum(\mathbf{x})$. The function inside the summation is just $f_i(\mathbf{x}) = x_i$ and the gradient is then:
\[\nabla y = \begin{bmatrix} \sum_i \frac{\partial f_i(\mathbf{x})}{\partial x_1},~ \sum_i \frac{\partial f_i(\mathbf{x})}{\partial x_2},~ \ldots,~ \sum_i \frac{\partial f_i(\mathbf{x})}{\partial x_n}  \end{bmatrix} = \begin{bmatrix} \sum_i \frac{\partial x_i}{\partial x_1},~ \sum_i \frac{\partial x_i}{\partial x_2},~ \ldots,~ \sum_i \frac{\partial x_i}{\partial x_n}  \end{bmatrix}\]
Because $\frac{\partial}{\partial x_j} x_i = 0$ for $j \neq i$, we can simplify to:
\[\nabla y = \begin{bmatrix} \frac{\partial x_1}{\partial x_1},~ \frac{\partial x_2}{\partial x_2},~ \ldots,~ \frac{\partial x_n}{\partial x_n}  \end{bmatrix} = \begin{bmatrix}1, 1, \ldots, 1\end{bmatrix} = \vec{1}^T\]
Notice that the result is a horizontal vector full of 1s, not a vertical vector, and so the gradient is $\vec{1}^T$.  (The $T$ exponent of $\vec{1}^T$ represents the transpose of the indicated vector. In this case, it flips a vertical vector to a horizontal vector.) It's very important to keep the shape of all of your vectors and matrices in order otherwise it's impossible to compute the derivatives of complex functions.

As another example, let's sum the result of multiplying a vector by a constant scalar.  If $y = sum(\mathbf{x} z)$ then $f_i(\mathbf{x},z) = x_i z$. The gradient is:
\[\begin{array}{lcl}
 \frac{\partial y}{\partial \mathbf{x}} & = & \begin{bmatrix} \sum_i \frac{\partial}{\partial x_1} x_i z,~ \sum_i \frac{\partial }{\partial x_2} x_i z,~ \ldots,~ \sum_i \frac{\partial}{\partial x_n} x_i z  \end{bmatrix}\\\\
 & = & \begin{bmatrix} \frac{\partial}{\partial x_1} x_1 z,~ \frac{\partial }{\partial x_2} x_2 z,~ \ldots,~ \frac{\partial}{\partial x_n} x_n z  \end{bmatrix}\\\\
 & = & \begin{bmatrix} z, z, \ldots, z \end{bmatrix}\\
\end{array}\]
The derivative with respect to scalar variable $z$ is $1 \times 1$:
\[\begin{array}{lcl}
 \frac{\partial y}{\partial z} & = & \frac{\partial}{\partial z} \sum_{i=1}^n x_i z\\\\
 & = & \sum_i \frac{\partial}{\partial z} x_i z\\\\
 & = & \sum_i x_i\\\\
 & = & sum(\mathbf{x})\\
\end{array}\]

\subsection{The Chain Rules}\label{sec4.5}

We can't compute partial derivatives of very complicated functions using just the basic matrix calculus rules we've seen so far.  For example, we can't take the derivative of nested expressions like $sum(\mathbf{w}+\mathbf{x})$ directly without reducing it to its scalar equivalent. We need to be able to combine our basic vector rules using what we can call the {\em vector chain rule}.   Unfortunately, there are a number of rules for differentiation that fall under the name ``chain rule'' so we have to be careful which chain rule we're talking about. Part of our goal here is to clearly define and name three different chain rules and indicate in which situation they are appropriate. To get warmed up, we'll start with what we'll call the {\em single-variable chain rule}, where we want the derivative of a scalar function with respect to a scalar. Then we'll move on to an important concept called the {\em total derivative} and use it to define what we'll pedantically call the {\em single-variable total-derivative chain rule}. Then, we'll be ready for the vector chain rule in its full glory as needed for neural networks.

The chain rule is conceptually a divide and conquer strategy (like Quicksort) that breaks complicated expressions into subexpressions whose derivatives are easier to compute.  Its power derives from the fact that we can process each simple subexpression in isolation yet still combine the  intermediate results to get the correct overall result.

The chain rule comes into play when we need the derivative of an expression composed of nested subexpressions. For example, we need the chain rule when confronted with expressions like $\frac{d}{dx} sin(x^2)$.  The outermost expression takes the $sin$ of an intermediate result, a nested subexpression that squares $x$. Specifically, we need the single-variable chain rule, so let's start by digging into that in more detail.

\subsubsection{Single-variable chain rule}\label{sec4.5.1}

Let's start with the solution to the derivative of our nested expression: $\frac{d}{dx} sin(x^2) = 2xcos(x^2)$.  It doesn't take a mathematical genius to recognize components of the solution that smack of scalar differentiation rules, $\frac{d}{dx}x^2 = 2x$ and $\frac{d}{du} sin(u) = cos(u)$. It looks like the solution is to multiply the derivative of the outer expression by the derivative of the inner expression or ``chain the pieces together,'' which is exactly right. In this section, we'll explore the general principle at work and provide a process that works for highly-nested expressions of a single variable.

Chain rules are typically defined in terms of nested functions, such as $y = f(g(x))$ for single-variable chain rules. (You will also see the chain rule defined using function composition $(f \circ g)(x)$, which is the same thing.)  Some sources write the derivative using shorthand notation $y' = f'(g(x))g'(x)$, but that hides the fact that we are introducing an intermediate variable: $u = g(x)$, which we'll see shortly. It's better to define the \href{http://m.wolframalpha.com/input/?i=chain+rule}{single-variable chain rule} of $f(g(x))$ explicitly so we never take the derivative with respect to the wrong variable. Here is the formulation of the single-variable chain rule we recommend:
\[\frac{dy}{dx} = \frac{dy}{du}\frac{du}{dx}\]
To deploy the single-variable chain rule, follow these steps:
\begin{enumerate}
\item Introduce intermediate variables for nested subexpressions and subexpressions for both binary and unary operators; e.g., $\times$ is binary, $sin(x)$ and other trigonometric functions are usually unary because there is a single operand. This step normalizes all equations to single operators or function applications.\item Compute derivatives of the intermediate variables with respect to their parameters.\item Combine all derivatives of intermediate variables by multiplying them together to get the overall result.\item Substitute intermediate variables back in if any are referenced in the derivative equation.
\end{enumerate}
The third step puts the ``chain'' in ``chain rule'' because it chains together intermediate results. Multiplying the intermediate derivatives together is the common theme among all variations of the chain rule.

Let's try  this process on $y = f(g(x)) = sin(x^2)$:
\begin{enumerate}
\item Introduce intermediate variables. Let $u = x^2$ represent subexpression $x^2$ (shorthand for $u(x) = x^2$). This gives us:
   
\[ \begin{array}{lllll}
  u &=& x^2 &&(\text{relative to definition }f(g(x)), g(x) = x^2)\\
	y &=& sin(u) && (y = f(u) = sin(u))
\end{array} \]
The order of these subexpressions does not affect the answer, but we recommend working in the reverse order of operations dictated by the nesting (innermost to outermost). That way, expressions and derivatives are always functions of previously-computed elements. 
\item Compute derivatives.\[ \begin{array}{lllll}
 \frac{du}{dx} &=& 2x && (\text{Take derivative with respect to }x)\\
	 \frac{dy}{du} &=& cos(u) && (\text{Take derivative with respect to }u \text{ not }x)
  \end{array} \]\item Combine.\[\frac{dy}{dx} = \frac{dy}{du} \frac{du}{dx} = cos(u)2x\]\item Substitute.\[\frac{dy}{dx} = \frac{dy}{du} \frac{du}{dx} = cos(x^2)2x = 2xcos(x^2)\]
\end{enumerate}
Notice how easy it is to compute the derivatives of the intermediate variables in isolation! The chain rule says it's legal to do that and tells us how to combine the intermediate results to get $2xcos(x^2)$.

You can think of the combining step of the chain rule in terms of units canceling. If we let $y$ be miles, $x$ be the gallons in a gas tank, and $u$ as gallons we can interpret $\frac{dy}{dx} = \frac{dy}{du} \frac{du}{dx}$ as $\frac{miles}{tank} = \frac{miles}{gallon} \frac{gallon}{tank}$. The $gallon$ denominator and numerator cancel.

Another way to to think about the single-variable chain rule is to visualize the overall expression as a dataflow diagram or chain of operations (or \href{https://en.wikipedia.org/wiki/Abstract\_syntax\_tree}{abstract syntax tree} for compiler people):

\begin{center}
\includegraphics{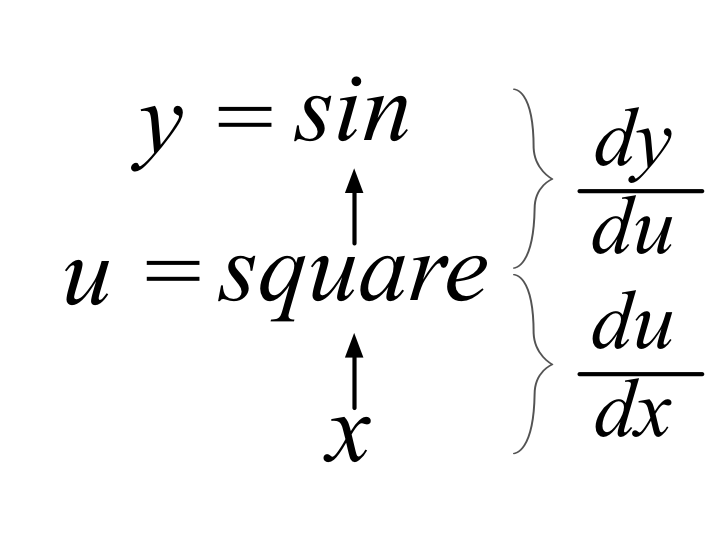}
\end{center}

Changes to function parameter $x$ bubble up through a squaring operation then through a $sin$ operation to change result $y$. You can think of $\frac{du}{dx}$ as ``getting changes from $x$ to $u$'' and $\frac{dy}{du}$ as ``getting changes from $u$ to $y$.'' Getting from $x$ to $y$ requires an intermediate hop. The chain rule is, by convention, usually written from the output variable down to the parameter(s), $\frac{dy}{dx} = \frac{dy}{du} \frac{du}{dx}$. But, the $x$-to-$y$ perspective would be more clear if we reversed the flow and used the equivalent $\frac{dy}{dx} = \frac{du}{dx}\frac{dy}{du}$.

{\bf Conditions under which the single-variable chain rule applies}. Notice that there is a single dataflow path from $x$ to the root $y$.  Changes in $x$ can influence output $y$ in only one way.  That is the condition under which we can apply the single-variable chain rule. An easier condition to remember, though one that's a bit looser, is that none of the intermediate subexpression functions, $u(x)$ and $y(u)$, have more than one parameter.  Consider $y(x) = x+x^2$, which would become $y(x,u) = x+u$ after introducing intermediate variable $u$.  As we'll see in the next section, $y(x,u)$ has multiple paths from $x$ to $y$. To handle that situation, we'll deploy the single-variable total-derivative chain rule.

As an aside for those interested in automatic differentiation, papers and library documentation use terminology {\em forward differentiation} and {\em backward differentiation} (for use in the back-propagation algorithm). From a dataflow perspective, we are computing a forward differentiation because it follows the normal data flow direction.  Backward differentiation, naturally, goes the other direction and we're asking how a change in the output would affect function parameter $x$. Because backward differentiation can determine changes in all function parameters at once, it turns out to be much more efficient for computing the derivative of functions with lots of parameters. Forward differentiation, on the other hand, must consider how a change in each parameter, in turn, affects the function output $y$. The following table emphasizes the order in which partial derivatives are computed for the two techniques.

\begin{tabular}{ll}
{\bf Forward differentiation from $x$ to $y$}&{\bf Backward differentiation from $y$ to $x$}\\

$\frac{dy}{dx} = \frac{du}{dx}\frac{dy}{du}$&$\frac{dy}{dx} = \frac{dy}{du} \frac{du}{dx}$\\

\end{tabular}

Automatic differentiation is beyond the scope of this article, but we're setting the stage for a future article.

Many readers can solve $\frac{d}{dx}sin(x^2)$ in their heads, but our goal is a process that will work even for  very complicated expressions. This process is also how \href{https://en.wikipedia.org/wiki/Automatic\_differentiation}{automatic differentiation} works in libraries like PyTorch. So, by solving derivatives manually  in this way, you're also learning how to define functions for custom neural networks in PyTorch.

With deeply nested expressions, it helps to think about deploying the chain rule the way a compiler unravels nested function calls like $f_4(f_3(f_2(f_1(x))))$ into a sequence (chain) of calls. The result of calling function $f_i$ is saved to a temporary variable called a register, which is then passed as a parameter to $f_{i+1}$.  Let's see how that looks in practice by using our process on a highly-nested equation like $y = f(x) = ln(sin(x^3)^2)$:
\begin{enumerate}
\item Introduce intermediate variables.\[\begin{array}{lllllllll}
 u_1 &=& f_1(x) &= x^3\\
 u_2 &= &f_2(u_1) &= sin(u_1)\\
 u_3 &= &f_3(u_2) &= u_2^2\\
 u_4 &=& f_4(u_3) &= ln(u_3) (y = u_4)
\end{array}\]\item  Compute derivatives.\[\begin{array}{lllllllll}
 \frac{d}{u_x} u_1 & = & \frac{d}{x} x^3 & = & 3x^2\\
 \frac{d}{u_1} u_2 & = & \frac{d}{u_1} sin(u_1) & = & cos(u_1) \\
 \frac{d}{u_2} u_3 & = & \frac{d}{u_2} u_2^2 & =& 2u_2\\
 \frac{d}{u_3} u_4 & = & \frac{d}{u_3} ln(u_3) & =& \frac{1}{u_3}\\
 \end{array}\]\item  Combine four intermediate values.\[\frac{dy}{dx} = \frac{d u_4}{dx} = \frac{d u_4}{du_3}\frac{du_3}{d u_2} \frac{du_2}{du_1} \frac{du_1}{dx} = \frac{1}{u_3}  2u_2  cos(u_1)  3x^2 = \frac{6u_2x^2cos(u_1)}{u_3}\]\item  Substitute.\[\frac{dy}{dx} = \frac{6sin(u_1)x^2cos(x^3)}{u_2^2} = \frac{6sin(x^3)x^2cos(x^3)}{sin(u_1)^2} = \frac{6sin(x^3)x^2cos(x^3)}{sin(x^3)^2} = \frac{6x^2cos(x^3)}{sin(x^3)}\]
\end{enumerate}
Here is a visualization of the data flow through the chain of operations from $x$ to $y$:

\begin{center}
\includegraphics{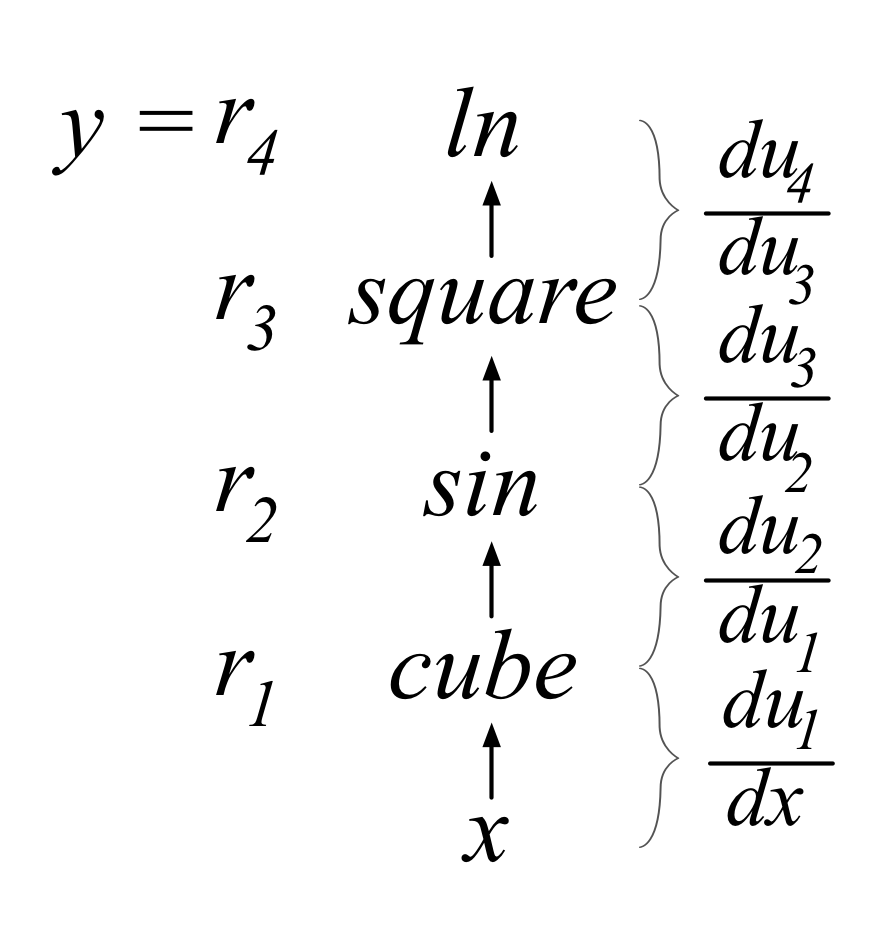}
\end{center}

At this point, we can handle derivatives of nested expressions of a single variable, $x$, using the chain rule but only if $x$ can affect $y$ through a single data flow path. To handle more complicated expressions, we need to extend our technique, which we'll do next.

\subsubsection{Single-variable total-derivative chain rule}\label{sec4.5.2}

Our single-variable chain rule has limited applicability because all intermediate variables must be functions of single variables. But, it demonstrates the core mechanism of the chain rule, that of multiplying out all derivatives of intermediate subexpressions. To handle more general expressions such as $y = f(x) = x+x^2$, however, we need to augment that basic chain rule.

Of course, we immediately see $\frac{dy}{dx} = \frac{d}{dx}x + \frac{d}{dx}x^2 = 1 + 2x$, but that is using the scalar  addition derivative rule, not the chain rule.  If we tried to apply the single-variable chain rule, we'd get the wrong answer. In fact, the previous chain rule is meaningless in this case because derivative operator $\frac{d}{dx}$ does not apply to multivariate functions, such as $u_2$ among our intermediate variables:
\[
\begin{array}{lllllllll}
 u_1(x) &=& x^2\\
 u_2(x,u_1) &=& x + u_1 & & & (y = f(x) = u_2(x,u_1))
\end{array}
\]
Let's try it anyway to see what happens. If we pretend that $\frac{du_2}{du_1} = 0 + 1 = 1$ and $\frac{du_1}{dx} = 2x$, then $\frac{dy}{dx} = \frac{du_2}{dx} = \frac{du_2}{du_1} \frac{du_1}{dx} = 2x$ instead of the right answer $1 + 2x$.  

Because $u_2(x,u) = x + u_1$ has multiple parameters, partial derivatives come into play. Let's blindly apply the partial derivative operator to all of our equations and see what we get:
\[
\begin{array}{lllllllll}
	\frac{\partial u_1(x)}{\partial x} &=& 2x &&&(\text{same as }\frac{du_1(x)}{dx})\\
	\frac{\partial u_2(x,u_1)}{\partial u_1} &=& \frac{\partial }{\partial u_1}(x + u_1) = 0 + 1 = 1\\
	\frac{\partial u_2(x,u_1)}{\partial x} &\includegraphics[scale=.012]{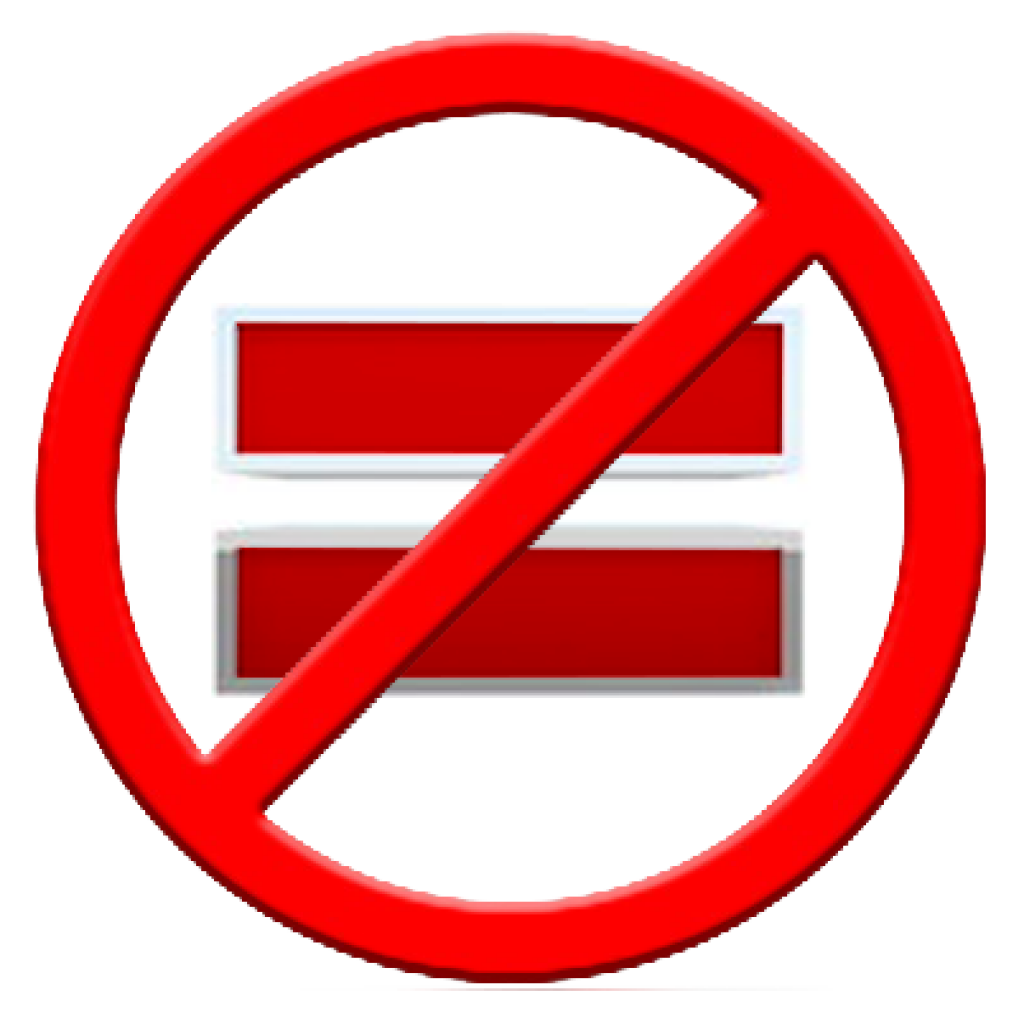}& \frac{\partial }{\partial x}(x + u_1) = 1 + 0 = 1 & & &(\text{something's not quite right here!})\\
\end{array}
\]
Ooops! The partial $\frac{\partial u_2(x,u_1)}{\partial x}$ is wrong because it violates a key assumption for partial derivatives. When taking the partial derivative with respect to $x$, the other variables must not vary as $x$ varies. Otherwise, we could not act as if the other variables were constants. Clearly, though, $u_1(x)=x^2$ is a function of $x$ and therefore varies with $x$. $\frac{\partial u_2(x,u_1)}{\partial x} \neq 1 + 0$ because $\frac{\partial u_1(x)}{\partial x} \neq 0$. A quick look at the data flow diagram for $y=u_2(x,u_1)$ shows multiple paths from $x$ to $y$, thus, making it clear we need to consider direct and indirect (through $u_1(x)$) dependencies on $x$:

\begin{center}
\includegraphics{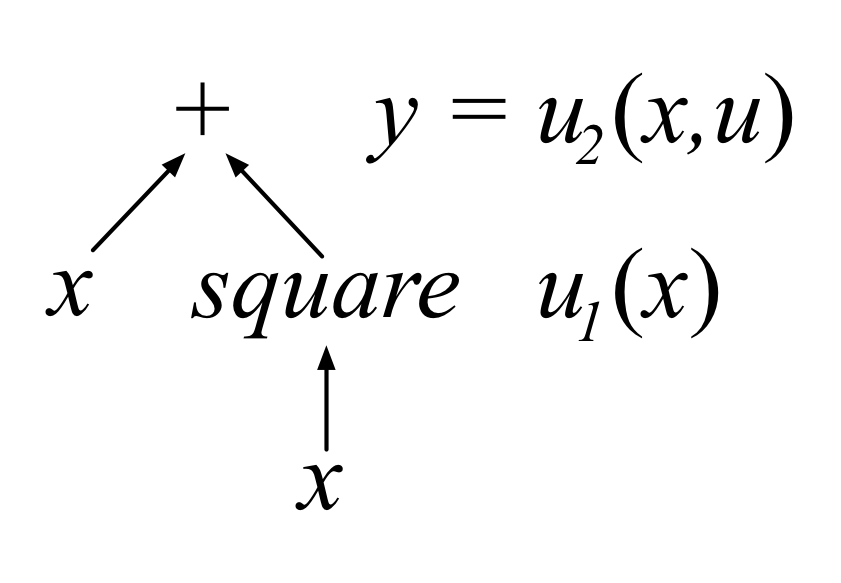}
\end{center}

A change in $x$ affects $y$ both as an operand of the addition and as the operand of the square operator. Here's an equation that describes how tweaks to $x$ affect the output:
\[\hat y = (x + \Delta x) + (x + \Delta x)^2\]
Then, $\Delta y = \hat y - y$, which we can read as ``the change in $y$ is the difference between the original $y$ and $y$ at a tweaked $x$.''

If we let $x=1$, then $y=1+1^2=2$. If we bump $x$ by 1, $\Delta x=1$, then $\hat y = (1+1) + (1+1)^2 = 2 + 4 = 6$. The change in $y$ is not $1$, as $\partial u_2 / u_1$ would lead us to believe, but $6-2 = 4$!

Enter the ``law'' of \href{https://en.wikipedia.org/wiki/Total\_derivative}{total derivatives}, which basically says that to compute $\frac{dy}{dx}$, we need to sum up all possible contributions from changes in $x$ to the change in $y$. The total derivative with respect to $x$ assumes all variables, such as $u_1$ in this case, are functions of $x$ and potentially vary as $x$ varies.   The total derivative of $f(x) = u_2(x,u_1)$ that depends on $x$ directly and indirectly via intermediate variable $u_1(x)$ is given by:
\[\frac{dy}{dx} = \frac{\partial f(x)}{\partial x} = \frac{\partial u_2(x,u_1)}{\partial x} = \frac{\partial u_2}{\partial x}\frac{\partial x}{\partial x} + \frac{\partial u_2}{\partial u_1}\frac{\partial u_1}{\partial x} = \frac{\partial u_2}{\partial x} + \frac{\partial u_2}{\partial u_1}\frac{\partial u_1}{\partial x}\]
Using this formula, we get the proper answer:
\[
\frac{dy}{dx} = \frac{\partial f(x)}{\partial x} = \frac{\partial u_2}{\partial x} + \frac{\partial u_2}{\partial u_1}\frac{\partial  u_1}{\partial  x} = 1 + 1 \times 2x = 1 + 2x
\]
That is an application of what we can call the {\em single-variable total-derivative chain rule}:
\[
\frac{\partial f(x,u_1,\ldots,u_n)}{\partial x} = \frac{\partial f}{\partial x} + \frac{\partial f}{\partial u_1}\frac{\partial  u_1}{\partial  x} + \frac{\partial f}{\partial u_2}\frac{\partial  u_2}{\partial  x} + \ldots + \frac{\partial f}{\partial u_n}\frac{\partial  u_n}{\partial x} = \frac{\partial f}{\partial x} + \sum_{i=1}^n \frac{\partial f}{\partial u_i}\frac{\partial  u_i}{\partial  x}
\]
The total derivative assumes all variables are potentially codependent whereas the partial derivative assumes all variables but $x$ are constants.

There is something subtle going on here with the notation. All of the derivatives are shown as partial derivatives because $f$ and $u_i$ are functions of multiple variables. This notation mirrors that of \href{http://mathworld.wolfram.com/TotalDerivative.html}{MathWorld's notation} but differs from  \href{https://en.wikipedia.org/wiki/Total\_derivative}{Wikipedia}, which uses ${d f(x,u_1,\ldots,u_n)}/{d x}$ instead (possibly to emphasize the total derivative nature of the equation). We'll stick with the partial derivative notation so that it's consistent with our discussion of the vector chain rule in the next section.

In practice, just keep in mind that when you take the total derivative with respect to $x$, other variables might also be functions of $x$ so add in their contributions as well.  The left side of the equation looks like a typical partial derivative but the right-hand side is actually the total derivative.  It's common, however, that many temporary variables are functions of a single parameter, which means that the single-variable total-derivative chain rule degenerates to the single-variable chain rule.

Let's look at a nested subexpression, such as $f(x) = sin(x + x^2)$.  We introduce three intermediate variables:
\[
\begin{array}{lllllllll}
	u_1(x) &=& x^2\\
	u_2(x,u_1) &=& x + u_1\\
	u_3(u_2) &=& sin(u_2) &&&(y = f(x) = u_3(u_2))
\end{array}
\]
and partials:
\[
\begin{array}{lllllllll}
	\frac{\partial u_1}{\partial x} &=& 2x\\
	\frac{\partial u_2}{\partial x} &=& \frac{\partial x}{\partial x} + \frac{\partial u_2}{\partial u_1}\frac{\partial u_1}{\partial x} &=& 1 + 1 \times 2x &=& 1+2x\\
	\frac{\partial f(x)}{\partial x} &=& \frac{\partial u_3}{\partial x} + \frac{\partial u_3}{\partial u_2}\frac{\partial u_2}{\partial x} &=& 0 + cos(u_2)\frac{\partial u_2}{\partial x} &=& cos(x+x^2)(1+2x)
\end{array}
\]
where both $\frac{\partial u_2}{\partial x}$ and $\frac{\partial f(x)}{\partial x}$ have $\frac{\partial u_i}{\partial x}$ terms that take into account the total derivative.

Also notice that the total derivative formula always {\bf sums} versus, say, multiplies terms $\frac{\partial f}{\partial u_i}\frac{\partial  u_i}{\partial  x}$.  It's tempting to think that summing up terms in the derivative makes sense because, for example, $y = x+x^2$ adds two terms. Nope. The total derivative is adding terms because it represents a weighted sum of all $x$ contributions to the change in $y$. For example, given $y = x \times x^2$ instead of $y = x + x^2$, the total-derivative chain rule formula still adds partial derivative terms. ($x \times x^2$  simplifies to $x^3$ but for this demonstration, let's not combine the terms.) Here are the intermediate variables and partial derivatives:
\[
\begin{array}{lllllllll}
 u_1(x) &=& x^2\\
 u_2(x,u_1) &=& x u_1 &&& (y = f(x) = u_2(x,u_1))\\
\\
 \frac{\partial u_1}{\partial x} &=& 2x\\
 \frac{\partial u_2}{\partial x} &=& u_1 &&&(\text{for } u_2 = x + u_1, \frac{\partial u_2}{\partial x} = 1)\\
 \frac{\partial u_2}{\partial u_1} &=& x &&&(\text{for }  u_2 = x + u_1, \frac{\partial u_2}{\partial u_1} = 1)
\end{array}
\]
The form of the total derivative remains the same, however:
\[\frac{dy}{dx} = \frac{\partial u_2}{\partial x} + \frac{\partial u_2}{\partial u_1}\frac{d u_1}{\partial  x} = u_1 + x 2x = x^2 + 2x^2 = 3x^2\]
It's the partials (weights) that change, not the formula, when the intermediate variable operators change.

Those readers with a strong calculus background might wonder why we aggressively introduce intermediate variables even for the non-nested subexpressions such as $x^2$ in $x+x^2$. We use this process for three reasons: (i) computing the derivatives for the simplified subexpressions is usually trivial, (ii) we can simplify the chain rule, and (iii) the process mirrors how automatic differentiation works in neural network libraries.

Using the intermediate variables even more aggressively, let's see how we can simplify our single-variable total-derivative chain rule to its final form. The goal is to get rid of the $\frac{\partial f}{\partial x}$ sticking out on the front like a sore thumb:
\[
\frac{\partial f(x,u_1,\ldots,u_n)}{\partial x} = \frac{\partial f}{\partial x} + \sum_{i=1}^n \frac{\partial f}{\partial u_i}\frac{\partial  u_i}{\partial  x}
\]
We can achieve that by simply introducing a new temporary variable as an alias for $x$: $u_{n+1} = x$. Then, the formula reduces to our final form:
\[
\frac{\partial f(u_1,\ldots,u_{n+1})}{\partial x} = \sum_{i=1}^{n+1} \frac{\partial f}{\partial u_i}\frac{\partial  u_i}{\partial  x}
\]
This chain rule that takes into consideration the total derivative degenerates to the single-variable chain rule when all intermediate variables are functions of a single variable.   Consequently, you can remember this more general formula to cover both cases.  As a bit of dramatic foreshadowing, notice that the summation sure looks like a vector dot product, $\frac{\partial f}{\partial \mathbf{u}} \cdot \frac{\partial \mathbf{u}}{\partial x}$, or  a vector multiply $\frac{\partial f}{\partial \mathbf{u}} \frac{\partial \mathbf{u}}{\partial x}$. 

Before we move on, a word of caution about terminology on the web. Unfortunately, the chain rule given in this section, based upon the total derivative, is universally called ``multivariable chain rule'' in calculus discussions, which is highly misleading! Only the intermediate variables are multivariate functions. The overall function, say, $f(x) = x + x^2$, is a scalar function that accepts a single parameter $x$. The derivative and parameter are scalars, not vectors, as one would expect with a so-called multivariate chain rule.  (Within the context of a non-matrix calculus class, ``multivariate chain rule'' is likely unambiguous.) To reduce confusion, we use ``single-variable total-derivative chain rule'' to spell out the distinguishing feature between the simple single-variable chain rule, $\frac{dy}{dx} = \frac{dy}{du}\frac{du}{dx}$, and this one.

\subsubsection{Vector chain rule}\label{sec4.5.3}

Now that we've got a good handle on the total-derivative chain rule, we're ready to tackle the chain rule for vectors of functions and vector variables. Surprisingly, this more general chain rule is just as simple looking as the single-variable chain rule for scalars. Rather than just presenting the vector chain rule, let's rediscover it ourselves so we get a firm grip on it. We can start by computing the derivative of a sample vector function with respect to a scalar, $\mathbf{y} = \mathbf{f}(x)$, to see if we can abstract a general formula.   
\[\begin{bmatrix}
	y_1(x)\\
	y_2(x)\\
\end{bmatrix} =
\begin{bmatrix}
	f_1(x)\\
	f_2(x)\\
\end{bmatrix} = 
\begin{bmatrix}
	ln(x^2)\\
	sin(3x)
\end{bmatrix}
\]
Let's introduce two intermediate variables, $g_1$ and $g_2$, one for each $f_i$ so that $y$ looks more like $\mathbf{y} = \mathbf{f}(\mathbf{g}(x))$:
\[ \begin{bmatrix}
	g_1(x)\\
	g_2(x)\\
\end{bmatrix} = \begin{bmatrix}
	x^2\\
	3x\\
\end{bmatrix}
\]\[ \begin{bmatrix}
	f_1(\mathbf{g})\\
	f_2(\mathbf{g})\\
\end{bmatrix} = \begin{bmatrix}
	ln(g_1)\\
	sin(g_2)\\
\end{bmatrix}\]
The derivative of vector $\mathbf{y}$ with respect to scalar $x$ is a vertical vector with elements computed using the single-variable total-derivative chain rule:
\[ \frac{\partial \mathbf{y}}{\partial x}  =
\begin{bmatrix}
	\frac{\partial f_1(\mathbf{g})}{\partial x}\\
	\frac{\partial f_2(\mathbf{g})}{\partial x}\\
\end{bmatrix} = \begin{bmatrix}
	\frac{\partial f_1}{\partial g_1}\frac{\partial g_1}{\partial x} + \frac{\partial f_1}{\partial g_2}\frac{\partial g_2}{\partial x}\\
	\frac{\partial f_2}{\partial g_1}\frac{\partial g_1}{\partial x} + \frac{\partial f_2}{\partial g_2}\frac{\partial g_2}{\partial x}\\
\end{bmatrix} = \begin{bmatrix}
	\frac{1}{g_1}2x + 0\\
	0 + cos(g_2)3\\
\end{bmatrix} = \begin{bmatrix}
	\frac{2x}{x^2}\\
	3cos(3x)\\
\end{bmatrix} = \begin{bmatrix}
	\frac{2}{x}\\
	3cos(3x)\\
\end{bmatrix}\]
Ok, so now we have the answer using just the scalar rules, albeit with the derivatives grouped into a vector. Let's try to abstract from that result what it looks like in vector form.  The goal is to convert the following vector of scalar operations to a vector operation. 
\[ \begin{bmatrix}
	\frac{\partial f_1}{\partial g_1}\frac{\partial g_1}{\partial x} + \frac{\partial f_1}{\partial g_2}\frac{\partial g_2}{\partial x}\\
	\frac{\partial f_2}{\partial g_1}\frac{\partial g_1}{\partial x} + \frac{\partial f_2}{\partial g_2}\frac{\partial g_2}{\partial x}\\
\end{bmatrix} \]
If we split the $\frac{\partial f_i}{\partial g_j}\frac{\partial g_j}{\partial x}$ terms, isolating the $\frac{\partial g_j}{\partial x}$ terms into a vector, we get a matrix by vector multiplication:
\[
\begin{bmatrix}
	\frac{\partial f_1}{\partial g_1} & \frac{\partial f_1}{\partial g_2}\\
	\frac{\partial f_2}{\partial g_1} & \frac{\partial f_2}{\partial g_2}\\
\end{bmatrix}\begin{bmatrix}
\frac{\partial g_1}{\partial x}\\
\frac{\partial g_2}{\partial x}\\
\end{bmatrix} = \frac{\partial \mathbf{f}}{\partial \mathbf{g}}\frac{\partial \mathbf{g}}{\partial x}
\]
That means that the Jacobian is the multiplication of two other Jacobians, which is kinda cool.  Let's check our results:
\[
\frac{\partial \mathbf{f}}{\partial \mathbf{g}}\frac{\partial \mathbf{g}}{\partial x} = \begin{bmatrix}
	\frac{1}{g_1} & 0\\
	0 & cos(g_2)\\
\end{bmatrix}\begin{bmatrix}
 2x\\
 3\\
\end{bmatrix} = \begin{bmatrix}
	\frac{1}{g_1}2x + 0\\
	0 + cos(g_2)3\\
\end{bmatrix} = \begin{bmatrix}
	\frac{2}{x}\\
	3cos(3x)\\
\end{bmatrix}
\]
Whew!  We get the same answer as the scalar approach. This vector chain rule for vectors of functions and a single parameter appears to be correct and, indeed, mirrors the single-variable chain rule. Compare the vector rule:
\[\frac{\partial}{\partial x} \mathbf{f}(\mathbf{g}(x)) = \frac{\partial \mathbf{f}}{\partial \mathbf{g}}\frac{\partial\mathbf{g}}{\partial x}\]
with the single-variable chain rule:
\[\frac{d}{dx} f(g(x)) = \frac{df}{dg}\frac{dg}{dx}\]
To make this formula work for multiple parameters or vector $\mathbf{x}$, we just have to change $x$ to vector $\mathbf{x}$ in the equation.  The effect is  that $\frac{\partial\mathbf{g}}{\partial \mathbf{x}}$ and the resulting Jacobian,  $\frac{\partial \mathbf{f}}{\partial \mathbf{x}}$, are now matrices instead of  vertical vectors. Our complete {\em vector chain rule} is:
\[
\begin{array}{lllllllll}
 \frac{\partial}{\partial \mathbf{x}} \mathbf{f}(\mathbf{g}(\mathbf{x})) = \frac{\partial \mathbf{f}}{\partial \mathbf{g}}\frac{\partial\mathbf{g}}{\partial \mathbf{x}} &(\text{Note: matrix multiply doesn't commute; order of }\frac{\partial \mathbf{f}}{\partial \mathbf{g}}\frac{\partial\mathbf{g}}{\partial \mathbf{x}} \text{ matters})\\
\end{array}
\]
The beauty of the vector formula over the single-variable chain rule is that it automatically takes into consideration the total derivative while maintaining the same notational simplicity.  The Jacobian contains all possible combinations of $f_i$ with respect to $g_j$ and $g_i$ with respect to $x_j$. For completeness, here are the two Jacobian components in their full glory:
\[
\frac{\partial}{\partial \mathbf{x}} \mathbf{f}(\mathbf{g}(\mathbf{x})) = \begin{bmatrix}
	\frac{\partial f_1}{\partial g_1} & \frac{\partial f_1}{\partial g_2} & \ldots & \frac{\partial f_1}{\partial g_k}\\
	\frac{\partial f_2}{\partial g_1} & \frac{\partial f_2}{\partial g_2} & \ldots & \frac{\partial f_2}{\partial g_k}\\
	& &\ldots\\
	\frac{\partial f_m}{\partial g_1} & \frac{\partial f_m}{\partial g_2} & \ldots & \frac{\partial f_m}{\partial g_k}\\
\end{bmatrix}\begin{bmatrix}
	\frac{\partial g_1}{\partial x_1} & \frac{\partial g_1}{\partial x_2} & \ldots & \frac{\partial g_1}{\partial x_n}\\
	\frac{\partial g_2}{\partial x_1} & \frac{\partial g_2}{\partial x_2} & \ldots & \frac{\partial g_2}{\partial x_n}\\
	& &\ldots\\
	\frac{\partial g_k}{\partial x_1} & \frac{\partial g_k}{\partial x_2} & \ldots & \frac{\partial g_k}{\partial x_n}\\
\end{bmatrix}
\]
where $m=|f|$, $n=|x|$, and $k=|g|$. The resulting Jacobian is $m \times n$ (an $m \times k$ matrix multiplied by a $k \times n$ matrix). 

Even within this $\frac{\partial \mathbf{f}}{\partial \mathbf{g}}\frac{\partial\mathbf{g}}{\partial \mathbf{x}}$ formula, we can simplify further because, for many applications, the Jacobians are square ($m=n$) and the off-diagonal entries are zero.  It is the nature of neural networks that the associated mathematics deals with functions of vectors not vectors of functions. For example, the neuron affine function has term $sum(\mathbf{w}\otimes\mathbf{x})$ and the activation function is $max(0,\mathbf{x})$; we'll consider derivatives of these functions in the next section.  

As we saw in a previous section, element-wise operations on vectors $\mathbf{w}$ and $\mathbf{x}$ yield diagonal matrices with elements $\frac{\partial w_i}{\partial x_i}$ because $w_i$ is a function purely of $x_i$ but not $x_j$ for $j \neq i$. The same thing happens here when $f_i$ is purely a function of $g_i$ and $g_i$ is purely a function of $x_i$:
\[\frac{\partial \mathbf{f}}{\partial \mathbf{g}} = diag(\frac{\partial f_i}{\partial g_i})\]\[\frac{\partial \mathbf{g}}{\partial \mathbf{x}} = diag(\frac{\partial g_i}{\partial x_i}) \]
In this situation, the vector chain rule simplifies to:
\[\frac{\partial}{\partial \mathbf{x}} \mathbf{f}(\mathbf{g}(\mathbf{x})) = diag(\frac{\partial f_i}{\partial g_i})diag(\frac{\partial g_i}{\partial x_i}) = diag(\frac{\partial f_i}{\partial g_i}\frac{\partial g_i}{\partial x_i})\]
Therefore, the Jacobian reduces to a diagonal matrix whose elements are the single-variable chain rule values.

After slogging through all of that mathematics, here's the payoff. All you need is the vector chain rule because the single-variable formulas are special cases of the vector chain rule. The following table summarizes the appropriate components to multiply in order to get the Jacobian.
\begin{center}

\begin{tabular}[t]{c|cccc}
  & 
\multicolumn{2}{c}{
  \begin{tabular}[t]{c}
  scalar\\
  \framebox(18,18){$x$}\\
  \end{tabular}} & &\begin{tabular}{c}
  vector\\
  \framebox(18,40){$\mathbf{x}$}\\
  \end{tabular} \\
  
  \begin{tabular}{c}$\frac{\partial}{\partial \mathbf{x}} \mathbf{f}(\mathbf{g}(\mathbf{x}))$
	   = $\frac{\partial \mathbf{f}}{\partial \mathbf{g}}\frac{\partial\mathbf{g}}{\partial \mathbf{x}}$
		\\
		\end{tabular} & \begin{tabular}[t]{c}
  scalar\\
  \framebox(18,18){$u$}\\
  \end{tabular} & \begin{tabular}{c}
  vector\\
  \framebox(18,40){$\mathbf{u}$}
  \end{tabular}& & \begin{tabular}{c}
  vector\\
  \framebox(18,40){$\mathbf{u}$}\\
  \end{tabular} \\
\hline
\\[\dimexpr-\normalbaselineskip+5pt]

\begin{tabular}[b]{c}
  scalar\\
  \framebox(18,18){$f$}\\
  \end{tabular} &\framebox(18,18){$\frac{\partial f}{\partial {u}}$} \framebox(18,18){$\frac{\partial u}{\partial {x}}$} ~~~& \raisebox{22pt}{\framebox(40,18){$\frac{\partial f}{\partial {\mathbf{u}}}$}} \framebox(18,40){$\frac{\partial \mathbf{u}}{\partial x}$} & ~~~&
\raisebox{22pt}{\framebox(40,18){$\frac{\partial f}{\partial {\mathbf{u}}}$}} \framebox(40,40){$\frac{\partial \mathbf{u}}{\partial \mathbf{x}}$}
\\
  
\begin{tabular}[b]{c}
  vector\\
  \framebox(18,40){$\mathbf{f}$}\\
  \end{tabular} & \framebox(18,40){$\frac{\partial \mathbf{f}}{\partial {u}}$} \raisebox{22pt}{\framebox(18,18){$\frac{\partial u}{\partial {x}}$}} & \framebox(40,40){$\frac{\partial \mathbf{f}}{\partial \mathbf{u}}$} \framebox(18,40){$\frac{\partial \mathbf{u}}{\partial x}$} & & \framebox(40,40){$\frac{\partial \mathbf{f}}{\partial \mathbf{u}}$} \framebox(40,40){$\frac{\partial \mathbf{u}}{\partial \mathbf{x}}$}\\
  
\end{tabular}

\end{center}

\section{The gradient of neuron activation}\label{sec5}

We now have all of the pieces needed to compute the derivative of a typical neuron activation for a single neural network computation unit with respect to the model parameters, $\mathbf{w}$ and $b$:
\[ activation(\mathbf{x}) = max(0, \mathbf{w} \cdot \mathbf{x} + b) \]
(This represents a neuron with fully connected weights and rectified linear unit activation. There are, however, other affine functions such as convolution and other activation functions, such as exponential linear units, that follow similar logic.)

Let's worry about $max$ later and focus on computing $\frac{\partial}{\partial \mathbf{w}} (\mathbf{w} \cdot \mathbf{x} + b)$ and $\frac{\partial}{\partial b} (\mathbf{w} \cdot \mathbf{x} + b)$. (Recall that neural networks learn through optimization of their weights and biases.)  We haven't discussed the derivative of the dot product yet, $y = \mathbf{f(w)} \cdot \mathbf{g(x)}$, but we can use the chain rule to avoid having to memorize yet another rule. (Note notation $y$ not $\mathbf{y}$ as the result is a scalar not a vector.) 

The dot product $\mathbf{w} \cdot \mathbf{x}$ is just the summation of the element-wise multiplication of the elements: $\sum_i^n (w_i x_i) = sum(\mathbf{w} \otimes \mathbf{x})$. (You might also find it useful to remember the linear algebra notation $\mathbf{w} \cdot \mathbf{x} = \mathbf{w}^{T} \mathbf{x}$.) We know how to compute the partial derivatives of $sum(\mathbf{x})$ and $\mathbf{w} \otimes \mathbf{x}$ but haven't looked at partial derivatives for $sum(\mathbf{w} \otimes \mathbf{x})$. We need the chain rule for that and so we can introduce an intermediate vector variable $\mathbf{u}$ just as we did using the single-variable chain rule:
\[
\begin{array}{lllllllll}
 \mathbf{u} &=& \mathbf{w} \otimes \mathbf{x}\\
 y &=& sum(\mathbf{u}) \\
\end{array}
\]
Once we've rephrased $y$, we recognize two subexpressions for which we already know the partial derivatives:
\[
\begin{array}{lllllllll}
 \frac{\partial  \mathbf{u}}{\partial \mathbf{w}} &=& \frac{\partial }{\partial \mathbf{w}} (\mathbf{w} \otimes \mathbf{x}) &=& diag(\mathbf{x})\\
 \frac{\partial y}{\partial \mathbf{u}} &=& \frac{\partial }{\partial \mathbf{u}} sum(\mathbf{u}) &=& \vec{1}^T\\
\end{array}
\]
The vector chain rule says to multiply the partials:
\[\frac{\partial y}{\partial \mathbf{w}} = \frac{\partial y}{\partial \mathbf{u}} \frac{\partial \mathbf{u}}{\partial \mathbf{w}} = \vec{1}^T  diag(\mathbf{x}) = \mathbf{x}^T\]
To check our results, we can grind the dot product down into a pure scalar function:
\[
\begin{array}{lllllllll}
 y &=& \mathbf{w} \cdot \mathbf{x} &=& \sum_i^n (w_i x_i)\\
 \frac{\partial y}{\partial w_j} &=& \frac{\partial}{\partial w_j} \sum_i (w_i x_i) &=& \sum_i \frac{\partial}{\partial w_j} (w_i x_i) &=& \frac{\partial}{\partial w_j} (w_j x_j) &=& x_j\\
\end{array}
\]
Then:
\[\frac{\partial y}{\partial \mathbf{w}} = [ x_1, \ldots, x_n ] = \mathbf{x}^T\]
Hooray! Our scalar results match the vector chain rule results. 

Now, let $y = \mathbf{w} \cdot \mathbf{x} + b$, the full expression within the $max$ activation function call. We have two different partials to compute, but we don't need the chain rule:
\[
\begin{array}{lllllllll}
 \frac{\partial y}{\partial \mathbf{w}} &=& \frac{\partial }{\partial \mathbf{w}}\mathbf{w} \cdot \mathbf{x} + \frac{\partial }{\partial \mathbf{w}}b &=& \mathbf{x}^T + \vec{0}^T &=& \mathbf{x}^T\\
 \frac{\partial y}{\partial b} &=& \frac{\partial }{\partial b}\mathbf{w} \cdot \mathbf{x} + \frac{\partial }{\partial b}b &=& 0 + 1 &=& 1\\
\end{array}
\]
Let's tackle the partials of the neuron activation, $max(0, \mathbf{w} \cdot \mathbf{x} + b)$. The use of the $max(0,z)$ function call on scalar $z$ just says to treat all negative $z$ values as 0.  The derivative of the max function is a piecewise function. When $z \leq 0$, the derivative is 0 because $z$ is a constant. When $z > 0$, the derivative of the max function is just the derivative of $z$, which is $1$:
\[ \frac{\partial}{\partial z}max(0,z) =
	\begin{cases}
	0 & z \leq 0\\
	\frac{dz}{dz}=1 & z > 0\\
\end{cases} \]
An aside on broadcasting functions across scalars. When one or both of the $max$ arguments are vectors, such as $max(0,\mathbf{x})$, we broadcast the single-variable function $max$ across the elements. This is an example of an element-wise unary operator.  Just to be clear:
\[
max(0,\mathbf{x}) = \begin{bmatrix}
 max(0,x_1)\\
 max(0,x_2)\\
 \ldots\\
 max(0,x_n)\\
\end{bmatrix}
\]
For the derivative of the broadcast version then, we get a vector of zeros and ones where:
\[
\frac{\partial}{\partial x_i}max(0,x_i) =
	\begin{cases}
	0 & x_i \leq 0\\
	\frac{dx_i}{dx_i}=1 & x_i > 0\\
\end{cases}\]\[\frac{\partial}{\partial \mathbf{x}}max(0,\mathbf{x}) =
\begin{bmatrix}
	\frac{\partial}{\partial x_1}max(0,x_1)\\
	\frac{\partial}{\partial x_2}max(0,x_2)\\
	\ldots\\
    \frac{\partial}{\partial x_n}max(0,x_n)
\end{bmatrix}\]
To get the derivative of the $activation(\mathbf{x})$ function, we need the chain rule because of the nested subexpression, $\mathbf{w} \cdot \mathbf{x} + b$. Following our process, let's introduce intermediate scalar variable $z$ to represent the affine function giving:
\[z(\mathbf{w},b,\mathbf{x}) = \mathbf{w} \cdot \mathbf{x} + b\]\[activation(z) = max(0,z)\]
The vector chain rule tells us:
\[\frac{\partial activation}{\partial \mathbf{w}} = \frac{\partial activation}{\partial z}\frac{\partial z}{\partial \mathbf{w}}\]
which we can rewrite as follows:
\[\frac{\partial activation}{\partial \mathbf{w}} = \begin{cases}
	0\frac{\partial z}{\partial \mathbf{w}}=\vec{0}^T & z \leq 0\\
	1\frac{\partial z}{\partial \mathbf{w}}=\frac{\partial z}{\partial \mathbf{w}} = \mathbf{x}^T & z > 0 ~~~(\text{we computed }\frac{\partial z}{\partial \mathbf{w}}=\mathbf{x}^T \text{ previously})\\
\end{cases}\]
and then substitute $z = \mathbf{w} \cdot \mathbf{x} + b$ back in:
\[\frac{\partial activation}{\partial \mathbf{w}} = \begin{cases}
	\vec{0}^T & \mathbf{w} \cdot \mathbf{x} + b \leq 0\\
	\mathbf{x}^T & \mathbf{w} \cdot \mathbf{x} + b > 0\\
\end{cases}\]
That equation matches our intuition.  When the activation function clips affine function output $z$ to 0, the derivative is zero with respect to any weight $w_i$. When $z > 0$, it's as if the $max$ function disappears and we get just the derivative of $z$ with respect to the weights. 

Turning now to the derivative of the neuron activation with respect to $b$, we get:
\[\frac{\partial activation}{\partial b} = \begin{cases}
	0\frac{\partial z}{\partial b} = 0 & \mathbf{w} \cdot \mathbf{x} + b \leq 0\\
	1\frac{\partial z}{\partial b} = 1 & \mathbf{w} \cdot \mathbf{x} + b > 0\\
\end{cases}\]
Let's use these partial derivatives now to handle the entire loss function.

\section{The gradient of the neural network loss function}\label{sec6}

Training a neuron requires that we take the derivative of our loss  or ``cost'' function with respect to the parameters of our model, $\mathbf{w}$ and $b$. Because we train with multiple vector inputs (e.g., multiple images) and scalar targets (e.g., one classification per image), we need some more notation. Let 
\[ X = [\mathbf{x}_1, \mathbf{x}_2, \ldots, \mathbf{x}_N]^T \]
where $N=|X|$, and then let 
\[ \mathbf{y} = [target(\mathbf{x}_1), target(\mathbf{x}_2), \ldots, target(\mathbf{x}_N)]^T \]
where $y_i$ is a scalar. Then the cost equation becomes:
\[ C(\mathbf{w},b,X,\mathbf{y}) = \frac{1}{N} \sum_{i=1}^{N} (y_i - activation(\mathbf{x}_i))^2 = \frac{1}{N} \sum_{i=1}^{N} (y_i - max(0, \mathbf{w}\cdot\mathbf{x}_i+b))^2 \]
Following our chain rule process introduces these intermediate variables:
\[
\begin{array}{lllllllll} 
 u(\mathbf{w},b,\mathbf{x}) &=& max(0, \mathbf{w}\cdot\mathbf{x}+b)\\
 v(y,u) &=& y - u\\
 C(v) &=& \frac{1}{N} \sum_{i=1}^N v^2\\
\end{array}
\]
Let's compute the gradient with respect to $\mathbf{w}$ first.

\subsection{The gradient with respect to the weights}\label{sec6.1}

From before, we know:
\[\frac{\partial }{\partial \mathbf{w}} u(\mathbf{w},b,\mathbf{x}) = \begin{cases}
	\vec{0}^T & \mathbf{w} \cdot \mathbf{x} + b \leq 0\\
	\mathbf{x}^T & \mathbf{w} \cdot \mathbf{x} + b > 0\\
\end{cases}\]
and
\[\frac{\partial v(y,u)}{\partial \mathbf{w}} = \frac{\partial}{\partial \mathbf{w}} (y - u) = \vec{0}^T - \frac{\partial u}{\partial \mathbf{w}} = -\frac{\partial u}{\partial \mathbf{w}} = \begin{cases}
	\vec{0}^T & \mathbf{w} \cdot \mathbf{x} + b \leq 0\\
	-\mathbf{x}^T & \mathbf{w} \cdot \mathbf{x} + b > 0\\
\end{cases}\]
Then, for the overall gradient, we get:
\begin{center}

\begin{eqnarray*}
 \frac{\partial C(v)}{\partial \mathbf{w}} & = & \frac{\partial }{\partial \mathbf{w}}\frac{1}{N} \sum_{i=1}^N v^2\\\\
 & = & \frac{1}{N} \sum_{i=1}^N \frac{\partial}{\partial \mathbf{w}} v^2\\\\
 & = & \frac{1}{N} \sum_{i=1}^N \frac{\partial v^2}{\partial v} \frac{\partial v}{\partial \mathbf{w}} \\\\
 & = & \frac{1}{N} \sum_{i=1}^N 2v \frac{\partial v}{\partial \mathbf{w}} \\\\
 & = & \frac{1}{N} \sum_{i=1}^N \begin{cases}
	2v\vec{0}^T = \vec{0}^T & \mathbf{w} \cdot \mathbf{x}_i + b \leq 0\\
	-2v\mathbf{x}^T & \mathbf{w} \cdot \mathbf{x}_i + b > 0\\
\end{cases}\\\\
 & = & \frac{1}{N} \sum_{i=1}^N \begin{cases}
	\vec{0}^T & \mathbf{w} \cdot \mathbf{x}_i + b \leq 0\\
	-2(y_i-u)\mathbf{x}_i^T & \mathbf{w} \cdot \mathbf{x}_i + b > 0\\
\end{cases}\\\\
 & = & \frac{1}{N} \sum_{i=1}^N \begin{cases}
	\vec{0}^T & \mathbf{w} \cdot \mathbf{x}_i + b \leq 0\\
	-2(y_i-max(0, \mathbf{w}\cdot\mathbf{x}_i+b))\mathbf{x}_i^T & \mathbf{w} \cdot \mathbf{x}_i + b > 0\\
\end{cases}\\
\phantom{\frac{\partial C(v)}{\partial \mathbf{w}}} & = & \frac{1}{N} \sum_{i=1}^N \begin{cases}
	\vec{0}^T & \mathbf{w} \cdot \mathbf{x}_i + b \leq 0\\
	-2(y_i-(\mathbf{w}\cdot\mathbf{x}_i+b))\mathbf{x}_i^T & \mathbf{w} \cdot \mathbf{x}_i + b > 0\\
\end{cases}\\\\
 & = & \begin{cases}
	\vec{0}^T & \mathbf{w} \cdot \mathbf{x}_i + b \leq 0\\
	\frac{-2}{N} \sum_{i=1}^N (y_i-(\mathbf{w}\cdot\mathbf{x}_i+b))\mathbf{x}_i^T & \mathbf{w} \cdot \mathbf{x}_i + b > 0\\
\end{cases}\\\\
 & = & \begin{cases}
	\vec{0}^T & \mathbf{w} \cdot \mathbf{x}_i + b \leq 0\\
	\frac{2}{N} \sum_{i=1}^N (\mathbf{w}\cdot\mathbf{x}_i+b-y_i)\mathbf{x}_i^T & \mathbf{w} \cdot \mathbf{x}_i + b > 0\\
\end{cases}
\end{eqnarray*}

\end{center}
To interpret that equation, we can substitute an error term $e_i = \mathbf{w}\cdot\mathbf{x}_i+b-y_i$ yielding:
\[
\frac{\partial C}{\partial \mathbf{w}} = \frac{2}{N} \sum_{i=1}^N e_i\mathbf{x}_i^T ~~~\text{(for the nonzero activation case)}
\]
From there, notice that this computation is a weighted average across all $\mathbf{x}_i$ in $X$. The weights are the error terms, the difference between the target output and the actual neuron output for each $\mathbf{x}_i$ input. The resulting gradient will, on average, point in the direction of higher cost or loss because large $e_i$ emphasize their associated $\mathbf{x}_i$. Imagine we only had one input vector, $N=|X|=1$, then the gradient is just $2e_1\mathbf{x}_1^T$.  If the error is 0, then the gradient is zero and we have arrived at the minimum loss. If $e_1$ is some small positive difference, the gradient is a small step in the direction of $\mathbf{x}_1$. If $e_1$ is large, the gradient is a large step in that direction. If $e_1$ is negative, the gradient is reversed, meaning the highest cost is in the negative direction.

Of course, we want to reduce, not increase, the loss, which is why the \href{https://en.wikipedia.org/wiki/Gradient\_descent}{gradient descent} recurrence relation takes the negative of the gradient to update the current position (for scalar learning rate  $\eta$):
\[ \mathbf{w}_{t+1} = \mathbf{w}_{t} - \eta \frac{\partial C}{\partial \mathbf{w}} \]
Because the gradient indicates the direction of higher cost, we want to update $\mathbf{x}$ in the opposite direction.

\subsection{The derivative with respect to the bias}\label{sec6.2}

To optimize the bias, $b$, we also need the partial with respect to $b$.  Here are the intermediate variables again:
\[
\begin{array}{lllllllll}
 u(\mathbf{w},b,\mathbf{x}) &=& max(0, \mathbf{w}\cdot\mathbf{x}+b)\\
 v(y,u) &=& y - u\\
 C(v) &=& \frac{1}{N} \sum_{i=1}^N v^2\\
\end{array}
\]
We computed the partial with respect to the bias for equation $u(\mathbf{w},b,\mathbf{x})$ previously:
\[\frac{\partial u}{\partial b} = \begin{cases}
	0 & \mathbf{w} \cdot \mathbf{x} + b \leq 0\\
	1 & \mathbf{w} \cdot \mathbf{x} + b > 0\\
\end{cases}\]
For $v$, the partial is:
\[\frac{\partial v(y,u)}{\partial b} = \frac{\partial}{\partial b} (y - u) = 0 - \frac{\partial u}{\partial b} = -\frac{\partial u}{\partial b} = \begin{cases}
	0 & \mathbf{w} \cdot \mathbf{x} + b \leq 0\\
	-1 & \mathbf{w} \cdot \mathbf{x} + b > 0\\
\end{cases}\]
And for the partial of the cost function itself we get:
\begin{center}

\begin{eqnarray*}
 \frac{\partial C(v)}{\partial b} & = & \frac{\partial }{\partial b}\frac{1}{N} \sum_{i=1}^N v^2\\\\
 & = & \frac{1}{N} \sum_{i=1}^N \frac{\partial}{\partial b} v^2\\\\
 & = & \frac{1}{N} \sum_{i=1}^N \frac{\partial v^2}{\partial v} \frac{\partial v}{\partial b} \\\\
 & = & \frac{1}{N} \sum_{i=1}^N 2v \frac{\partial v}{\partial b} \\\\
 & = & \frac{1}{N} \sum_{i=1}^N \begin{cases}
 	0 & \mathbf{w} \cdot \mathbf{x} + b \leq 0\\
 	-2v & \mathbf{w} \cdot \mathbf{x} + b > 0\\
 \end{cases}\\\\
 & = & \frac{1}{N} \sum_{i=1}^N \begin{cases}
 	0 & \mathbf{w} \cdot \mathbf{x} + b \leq 0\\
 	-2(y_i-max(0, \mathbf{w}\cdot\mathbf{x}_i+b)) & \mathbf{w} \cdot \mathbf{x} + b > 0\\
 \end{cases}\\\\
 & = & \frac{1}{N} \sum_{i=1}^N \begin{cases}
 	0 & \mathbf{w} \cdot \mathbf{x} + b \leq 0\\
 	2(\mathbf{w}\cdot\mathbf{x}_i+b-y_i) & \mathbf{w} \cdot \mathbf{x} + b > 0\\
 \end{cases}\\\\
 & = & \begin{cases}
	0 & \mathbf{w} \cdot \mathbf{x}_i + b \leq 0\\
	\frac{2}{N} \sum_{i=1}^N (\mathbf{w}\cdot\mathbf{x}_i+b-y_i) & \mathbf{w} \cdot \mathbf{x}_i + b > 0\\
 \end{cases}
\end{eqnarray*}

\end{center}
As before, we can substitute an error term:
\[ \frac{\partial C}{\partial b} = \frac{2}{N} \sum_{i=1}^N e_i ~~~\text{(for the nonzero activation case)} \]
The partial derivative is then just the average error or zero, according to the activation level. To update the neuron bias, we nudge it in the opposite direction of increased cost:
\[ b_{t+1} = b_{t} - \eta \frac{\partial C}{\partial b} \]
In practice, it is convenient to combine $\mathbf{w}$ and $b$ into a single vector parameter rather than having to deal with two different partials: $\hat{\mathbf{w}} = [\mathbf{w}^T, b]^T$. This requires a tweak to the input vector $\mathbf{x}$ as well but simplifies the activation function. By tacking a 1 onto the end of $\mathbf{x}$, $\hat{\mathbf{x}} = [\mathbf{x}^T,1]$, $\mathbf{w} \cdot \mathbf{x} + b$ becomes $\hat{\mathbf{w}} \cdot \hat{\mathbf{x}}$.  

This finishes off the optimization of the neural network loss function because we have the two partials necessary to perform a gradient descent.

\section{Summary}\label{sec7}

Hopefully you've made it all the way through to this point.  You're well on your way to understanding matrix calculus!  We've included a reference that summarizes all of the rules from this article in the next section. Also check out the annotated resource link below.

Your next step would be to learn about the partial derivatives of matrices not just vectors. For example, you can take a look at the matrix differentiation section of \href{https://atmos.washington.edu/~dennis/MatrixCalculus.pdf}{Matrix calculus}.  

{\bf Acknowledgements}. We thank \href{https://www.usfca.edu/faculty/yannet-interian}{Yannet Interian} (Faculty in MS data science program at University of San Francisco) and \href{http://www.cs.usfca.edu/~duminsky/}{David Uminsky} (Faculty/director of MS data science) for their help with the notation presented here.

\section{Matrix Calculus Reference}\label{reference}

\subsection{Gradients and Jacobians}\label{sec8.1}

The {\em gradient} of a function of two variables is a horizontal 2-vector:
\[
\nabla f(x,y)  = [ \frac{\partial f(x,y)}{\partial x}, \frac{\partial f(x,y)}{\partial y}]
\]
The {\em Jacobian} of a vector-valued function that is a function of a vector is an $m \times n$ ($m=|\mathbf{f}|$ and $n=|\mathbf{x}|$) matrix containing all possible scalar partial derivatives:
\[
\frac{\partial \mathbf{y}}{\partial \mathbf{x}} = \begin{bmatrix}
\nabla f_1(\mathbf{x}) \\
\nabla f_2(\mathbf{x})\\
\ldots\\
\nabla f_m(\mathbf{x})
\end{bmatrix} = \begin{bmatrix}
\frac{\partial}{\partial \mathbf{x}} f_1(\mathbf{x}) \\
\frac{\partial}{\partial \mathbf{x}} f_2(\mathbf{x})\\
\ldots\\
\frac{\partial}{\partial \mathbf{x}} f_m(\mathbf{x})
\end{bmatrix} = \begin{bmatrix}
\frac{\partial}{\partial {x_1}} f_1(\mathbf{x}) \frac{\partial}{\partial {x_2}} f_1(\mathbf{x}) \ldots \frac{\partial}{\partial {x_n}} f_1(\mathbf{x}) \\
\frac{\partial}{\partial {x_1}} f_2(\mathbf{x}) \frac{\partial}{\partial {x_2}} f_2(\mathbf{x}) \ldots \frac{\partial}{\partial {x_n}} f_2(\mathbf{x}) \\
\ldots\\
\frac{\partial}{\partial {x_1}} f_m(\mathbf{x}) \frac{\partial}{\partial {x_2}} f_m(\mathbf{x}) \ldots \frac{\partial}{\partial {x_n}} f_m(\mathbf{x}) \\
\end{bmatrix}
\]
The Jacobian of the identity function $\mathbf{f(x)} = \mathbf{x}$ is $I$.

\subsection{Element-wise operations on vectors}\label{sec8.2}

Define generic {\em element-wise operations} on vectors $\mathbf{w}$ and $\mathbf{x}$ using operator $\bigcirc$ such as $+$:
\[
\begin{bmatrix}
           y_1\\
           y_2\\
           \vdots \\
           y_n\\
           \end{bmatrix} = \begin{bmatrix}
           f_{1}(\mathbf{w}) \bigcirc g_{1}(\mathbf{x})\\
           f_{n}(\mathbf{w}) \bigcirc g_{2}(\mathbf{x})\\
           \vdots \\
           f_{n}(\mathbf{w}) \bigcirc g_{n}(\mathbf{x})\\
         \end{bmatrix}\]
The Jacobian with respect to $\mathbf{w}$ (similar for $\mathbf{x}$) is:
\[ J_\mathbf{w} = 
\frac{\partial \mathbf{y}}{\partial \mathbf{w}}  = \begin{bmatrix}
\frac{\partial}{\partial w_1} ( f_{1}(\mathbf{w}) \bigcirc g_{1}(\mathbf{x}) ) & \frac{\partial}{\partial w_2} ( f_{1}(\mathbf{w}) \bigcirc g_{1}(\mathbf{x}) ) & \ldots & \frac{\partial}{\partial w_n} ( f_{1}(\mathbf{w}) \bigcirc g_{1}(\mathbf{x}) )\\
\frac{\partial}{\partial w_1} ( f_{2}(\mathbf{w}) \bigcirc g_{2}(\mathbf{x}) ) & \frac{\partial}{\partial w_2} ( f_{2}(\mathbf{w}) \bigcirc g_{2}(\mathbf{x}) ) & \ldots & \frac{\partial}{\partial w_n} ( f_{2}(\mathbf{w}) \bigcirc g_{2}(\mathbf{x}) )\\
& \ldots\\
\frac{\partial}{\partial w_1} ( f_{n}(\mathbf{w}) \bigcirc g_{n}(\mathbf{x}) ) & \frac{\partial}{\partial w_2} ( f_{n}(\mathbf{w}) \bigcirc g_{n}(\mathbf{x}) ) & \ldots & \frac{\partial}{\partial w_n} ( f_{n}(\mathbf{w}) \bigcirc g_{n}(\mathbf{x}) )\\
\end{bmatrix} \]
Given the constraint ({\em element-wise diagonal condition}) that $f_i(\mathbf{w})$ and $g_i(\mathbf{x})$ access at most $w_i$ and $x_i$, respectively, the Jacobian simplifies to a diagonal matrix:
\[
\frac{\partial \mathbf{y}}{\partial \mathbf{w}} = diag \left( \frac{\partial}{\partial w_1}(f_{1}(w_1) \bigcirc g_{1}(x_1)),~ \frac{\partial}{\partial w_2}(f_{2}(w_2) \bigcirc g_{2}(x_2)),~ \ldots,~ \frac{\partial}{\partial w_n}(f_{n}(w_n) \bigcirc g_{n}(x_n)) \right)
\]
Here are some sample element-wise  operators:
\[
\begin{array}{lll}
 \text{\bf Op} & \text{\bf Partial with respect to } \mathbf{w} & \text{\bf Partial with respect to }\mathbf{x}\\
 + & \frac{\partial (\mathbf{w+x})}{\partial \mathbf{w}} = I & \frac{\partial (\mathbf{w+x})}{\partial \mathbf{x}} =  I\\
 - & \frac{\partial (\mathbf{w-x})}{\partial \mathbf{w}}  = I & \frac{\partial (\mathbf{w-x})}{\partial \mathbf{x}}  = -I \\
 \otimes & \frac{\partial (\mathbf{w \otimes x})}{\partial \mathbf{w}}  =  diag(\mathbf{x}) & \frac{\partial (\mathbf{w \otimes x})}{\partial \mathbf{x}}  =  diag(\mathbf{w})\\
 \oslash & \frac{\partial (\mathbf{w \oslash x})}{\partial \mathbf{w}}  =  diag(\ldots \frac{1}{x_i} \ldots) & \frac{\partial (\mathbf{w \oslash x})}{\partial \mathbf{x}}  =  diag(\ldots \frac{-w_i}{x_i^2} \ldots)\\
\end{array}
\]

\subsection{Scalar expansion}\label{sec8.3}

Adding scalar $z$  to vector $\mathbf{x}$, $\mathbf{y} = \mathbf{x} + z$, is really $\mathbf{y} = \mathbf{f(x)} + \mathbf{g}(z)$ where $\mathbf{f(x)} = \mathbf{x}$ and $\mathbf{g}(z) = \vec{1} z$.
\[
\frac{\partial}{\partial \mathbf{x}} ( \mathbf{x} + z ) = diag(\vec{1}) = I
\]\[
\frac{\partial}{\partial z} ( \mathbf{x} + z ) = \vec{1}
\]
Scalar multiplication yields:
\[\frac{\partial}{\partial \mathbf{x}} ( \mathbf{x} z ) = I z\]\[\frac{\partial}{\partial z} ( \mathbf{x} z ) = \mathbf{x}\]

\subsection{Vector reductions}\label{sec8.4}

The partial derivative of a vector sum with respect to one of the vectors is:
\[
\nabla_{\mathbf{x}} y = \begin{array}{lcl}
 \frac{\partial y}{\partial \mathbf{x}} & = & \begin{bmatrix} \frac{\partial y}{\partial x_1}, \frac{\partial y}{\partial x_2}, \ldots, \frac{\partial y}{\partial x_n} \end{bmatrix} = \begin{bmatrix} \sum_i \frac{\partial f_i(\mathbf{x})}{\partial x_1},~ \sum_i \frac{\partial f_i(\mathbf{x})}{\partial x_2},~ \ldots,~ \sum_i \frac{\partial f_i(\mathbf{x})}{\partial x_n}  \end{bmatrix}\\
\end{array}
\]
For $y = sum(\mathbf{x})$:
\[\nabla_\mathbf{x} y = \vec{1}^T\]
For $y = sum(\mathbf{x}z)$ and $n = |x|$, we get:
\[\nabla_\mathbf{x} y = [z, z, \ldots, z]\]\[\nabla_z y = sum(\mathbf{x}) \]
Vector dot product $y = \mathbf{f(w)} \cdot \mathbf{g(x)} = \sum_i^n (w_i x_i) = sum(\mathbf{w} \otimes \mathbf{x})$.  Substituting $\mathbf{u} = \mathbf{w} \otimes \mathbf{x}$ and using the vector chain rule, we get:
\[
\begin{array}{lcl}
 \frac{d \mathbf{u}}{d\mathbf{x}} = \frac{d}{d\mathbf{x}} (\mathbf{w} \otimes \mathbf{x}) = diag(\mathbf{w})\\
 \frac{dy}{d\mathbf{u}} = \frac{d}{d\mathbf{u}} sum(\mathbf{u}) = \vec{1}^T\\
 \frac{dy}{d\mathbf{x}} = \frac{dy}{d\mathbf{u}} \times \frac{d\mathbf{u}}{d\mathbf{x}} = \vec{1}^T \times diag(\mathbf{w}) = \mathbf{w}^T
\end{array}
\]
Similarly, $\frac{dy}{d\mathbf{w}} = \mathbf{x}^T$.

\subsection{Chain rules}\label{sec8.5}

The {\em vector chain rule} is the general form as it degenerates to the others. When $f$ is a function of a single variable $x$ and all intermediate variables $u$ are functions of a single variable, the single-variable chain rule applies. When some or all of the intermediate variables are functions of multiple variables, the single-variable total-derivative chain rule applies. In all other cases, the vector chain rule applies.

\begin{tabular}{lll}
{\bf Single-variable rule}&{\bf Single-variable total-derivative rule}&{\bf Vector  rule}\\

$\frac{df}{dx} = \frac{df}{du}\frac{du}{dx}$&$\frac{\partial f(u_1,\ldots,u_n)}{\partial x} = \frac{\partial f}{\partial \mathbf{u}} \frac{\partial \mathbf{u}}{\partial x}$&$\frac{\partial}{\partial \mathbf{x}} \mathbf{f}(\mathbf{g}(\mathbf{x})) = \frac{\partial \mathbf{f}}{\partial \mathbf{g}}\frac{\partial\mathbf{g}}{\partial \mathbf{x}}$\\

\end{tabular}

\section{Notation}\label{notation}

Lowercase letters in bold font such as $\mathbf{x}$ are vectors and those in italics font like $x$ are scalars. $x_i$ is the $i^{th}$ element of vector $\mathbf{x}$ and is in italics because a single vector element is a scalar. $|\mathbf{x}|$ means ``length of vector $\mathbf{x}$.''

The $T$ exponent of $\mathbf{x}^T$ represents the transpose of the indicated vector.

$\sum_{i=a}^b x_i$ is just a for-loop that iterates $i$ from $a$ to $b$, summing all the $x_i$.

Notation $f(x)$ refers to a function called $f$ with an argument of $x$.

$I$ represents the square ``identity matrix'' of appropriate dimensions that is zero everywhere but the diagonal, which contains all ones.

$diag(\mathbf{x})$  constructs a matrix whose diagonal elements are taken from vector $\mathbf{x}$.

The dot product $\mathbf{w} \cdot \mathbf{x}$ is the summation of the element-wise multiplication of the elements: $\sum_i^n (w_i x_i) = sum(\mathbf{w} \otimes \mathbf{x})$. Or, you can look at it  as $w^T x$.

Differentiation $\frac{d}{dx}$ is an operator that maps a function of one parameter to another function.  That means that $\frac{d}{dx} f(x)$ maps $f(x)$ to its derivative with respect to $x$, which is the same thing as $\frac{df(x)}{dx}$. Also, if $y = f(x)$, then $\frac{dy}{dx} = \frac{df(x)}{dx} = \frac{d}{dx}f(x)$.

The partial derivative of the function with respect to $x$, $\frac{\partial}{\partial x} f(x)$, performs the usual scalar derivative holding all other variables constant.

The gradient of $f$ with respect to vector $\mathbf{x}$, $\nabla f(\mathbf{x})$, organizes all of the partial derivatives for a specific scalar function.

The Jacobian organizes the gradients of multiple functions into a matrix by stacking them:
\[    
J = \begin{bmatrix} 
	\nabla f_1(\mathbf{x})\\
	\nabla f_2(\mathbf{x})
\end{bmatrix}
\]
The following notation means that $y$ has the value $a$ upon $condition_1$ and value $b$ upon $condition_2$.
\[     
y = \begin{cases} 
	a & condition_1\\
	b & condition_2\\
\end{cases}
\]

\section{Resources}\label{sec10}

\href{http://www.wolframalpha.com/input/?i=D%5B%7Bx%5E2,+x%5E3%7D.%7B%7B1,2%7D,%7B3,4%7D%7D.%7Bx%5E2,+x%5E3%7D,+x%5D}{Wolfram Alpha} can do symbolic matrix algebra and there is also a cool dedicated \href{http://www.matrixcalculus.org/}{matrix calculus differentiator}.

When looking for resources on the web, search for ``matrix calculus'' not ``vector calculus.''  Here are some comments on the top links that come up from a \href{https://www.google.com/search?q=matrix+calculus\&oq=matrix+calculus}{Google search}:
\begin{itemize}
\item \href{https://en.wikipedia.org/wiki/Matrix\_calculus}{https://en.wikipedia.org/wiki/Matrix\_calculus}
The Wikipedia entry is actually quite good and they have a good description of the different layout conventions. Recall that we use the numerator layout where the variables go horizontally and the functions go vertically in the Jacobian. Wikipedia also has a good description of \href{https://en.wikipedia.org/wiki/Total\_derivative}{total derivatives}, but be careful that they use slightly different notation than we do. We always use the $\partial x$ notation not $dx$.
\item \href{http://www.ee.ic.ac.uk/hp/staff/dmb/matrix/calculus.html}{http://www.ee.ic.ac.uk/hp/staff/dmb/matrix/calculus.html}
This page has a section on matrix differentiation with some useful identities; this person uses numerator layout. This might be a good place to start after reading this article to learn about matrix versus vector differentiation.
\item \href{https://www.colorado.edu/engineering/CAS/courses.d/IFEM.d/IFEM.AppC.d/IFEM.AppC.pdf}{https://www.colorado.edu/engineering/CAS/courses.d/IFEM.d/IFEM.AppC.d/IFEM.AppC.pdf}
This is part of the course notes for ``Introduction to Finite Element Methods'' I believe by \href{https://www.colorado.edu/engineering/CAS/courses.d/IFEM.d}{Carlos A. Felippa}.  His Jacobians are transposed from our notation because he uses denominator layout.
\item \href{http://www.ee.ic.ac.uk/hp/staff/dmb/matrix/calculus.html}{http://www.ee.ic.ac.uk/hp/staff/dmb/matrix/calculus.html}
This page has a huge number of useful derivatives computed for a variety of vectors and matrices. A great cheat sheet. There is no discussion to speak of, just a set of rules.
\item \href{https://www.math.uwaterloo.ca/~hwolkowi/matrixcookbook.pdf}{https://www.math.uwaterloo.ca/\textasciitilde{}hwolkowi/matrixcookbook.pdf}
Another cheat sheet that focuses on matrix operations in general with more discussion than the previous item.
\item \href{https://www.comp.nus.edu.sg/~cs5240/lecture/matrix-differentiation.pdf}{https://www.comp.nus.edu.sg/\textasciitilde{}cs5240/lecture/matrix-differentiation.pdf}
A useful set of slides.

\end{itemize}
To learn more about neural networks and the mathematics behind optimization and back propagation, we highly recommend \href{http://neuralnetworksanddeeplearning.com/chap1.html}{Michael Nielsen's book}.

For those interested specifically in convolutional neural networks, check out \href{https://arxiv.org/pdf/1603.07285.pdf}{A guide to convolution arithmetic for deep learning}.

We reference the law of \href{https://en.wikipedia.org/wiki/Total\_derivative}{total derivative}, which is an important concept that just means derivatives with respect to $x$ must take into consideration the derivative with respect $x$ of all variables that are a function of $x$.

\end{document}